\RequirePackage{fix-cm}

\documentclass[twocolumn, natbib]{svjour3}

\smartqed  

\usepackage[ngerman,english]{babel}

\usepackage{graphicx}
\usepackage{multirow}
\usepackage{amsmath,amssymb,amsfonts}
\usepackage{mathrsfs}
\usepackage[title]{appendix}
\usepackage{xcolor}
\usepackage{textcomp}
\usepackage{manyfoot}
\usepackage{booktabs}
\usepackage{siunitx}%
\usepackage[normalem]{ulem}
\usepackage{array}
\usepackage{algorithm}
\usepackage{algorithmicx}
\usepackage{algpseudocode}
\usepackage{listings}
\usepackage{url}
\usepackage[hidelinks]{hyperref} 
\usepackage{microtype} 
\definecolor{1024}{RGB}{231,41,138}
\definecolor{2048}{RGB}{217,95,2}
\definecolor{4096}{RGB}{117,112,179}
\definecolor{full}{RGB}{27,158,119}

\newcommand{\tcb}[2]{\textcolor{#1}{\textbf{#2}}}

\makeatletter
\def\makeheadbox{}
\makeatother

\begin{document}

\title{Accuracy potential of visual localization exploiting high-end street-level imagery}

\author{
Jonas Meyer\ \and
Stephan Nebiker \and
Pascal Theiler \and
Norbert Haala
}


\institute{J. Meyer \and S. Nebiker \at
              Institute of Geomatics, FHNW University of Applied Sciences and Arts Northwestern Switzerland, 4132 Muttenz, Switzerland\\
              J. Meyer (corresponding author): \email{jonas.meyer@fhnw.ch}\\
              S. Nebiker: \email{stephan.nebiker@fhnw.ch}\\
           \and
           J. Meyer \and N. Haala \at
              Institute for Photogrammetry and Geoinformatics, University of Stuttgart, 70174 Stuttgart, Germany \\
              N. Haala: \email{norbert.haala@ifp.uni-stuttgart.de} \\
            \and
            P. Theiler \at
            iNovitas AG, 5400 Baden, Switzerland \\
            P. Theiler: \email{pascal.theiler@inovitas.ch}
}

\date{}

\maketitle

\begin{abstract}
Many applications, such as surveying, autonomous navigation, robotics and augmented reality, require accurate pose information with respect to a reference frame. Visual localization can complement GNSS positioning, whose availability and accuracy are often limited. However, the accuracy potential of visual localization has not yet been systematically investigated against survey-grade requirements, and no publicly available large-scale outdoor dataset provides ground-truth poses accurate enough for such an investigation. We address both gaps. First, we present the \textit{FHNW Muttenz} dataset. It covers a contiguous 10\,km street network, mapped in two mobile mapping campaigns about 1.5 years apart, and comprises high-resolution reference imagery as well as query sequences from four cameras in five scenes. The two campaigns are co-registered to each other with an accuracy of 5\,mm. Co-registration of the query images to the reference imagery yields ground-truth poses accurate to 6--9\,mm in position and well below 0.1$^\circ$ in orientation. All images are georeferenced with an accuracy of about 2.5\,cm. Second, we use this dataset to systematically evaluate sparse feature-based visual localization with a modular pipeline built from established components, using the high-resolution reference imagery directly as the scene representation. For daytime imagery and pre-calibrated cameras, the median pose errors with respect to the reference imagery are 1--5\,cm in translation and 0.03--0.1$^\circ$ in rotation. Given an accurately georeferenced scene representation, visual localization not only approaches the accuracy of survey-grade GNSS positioning but also provides full orientation information. The dataset is publicly available at \url{https://fhnw-muttenz-vl-dataset.github.io/}.


\end{abstract}

\section{Introduction}\label{introduction}
Estimating the 6 degree-of-freedom (DoF) pose of a sensor with respect to a reference frame is a prerequisite for many applications, including (autonomous) navigation, surveying, robotics, and augmented reality (AR). Typically, GNSS sensors are used in combination with IMUs for this task. Since this procedure heavily relies on good GNSS reception, it is applicable only outdoors. Even there, occlusions and multi-path effects significantly limit the applicability of GNSS in complex urban environments. Visual localization, which estimates the 6-DoF pose of an image with respect to a previously mapped scene, provides a complementary positioning modality: objects that occlude GNSS signals typically exhibit a high density of salient visual features.

Over the last two decades, many visual localization methods have been proposed, which can be divided into 2D image-based and 3D geometry-based approaches. 2D image-based methods represent the scene as a database of oriented images, from which image retrieval selects the reference images most similar to the query \citep{sivic_video_2003}. 3D geometry-based methods, in contrast, store the scene geometry in a globally consistent manner. This allows them to establish 2D--3D correspondences and to estimate the query pose with a Perspective-n-Point (PnP) solver \citep{haralick_review_1994} within a RANSAC framework \citep{fischler_random_1981}. Because no 3D geometry has to be computed or stored, 2D image-based approaches are more scalable and flexible, whereas 3D geometry-based approaches benefit from more extensive geometric reasoning and generally achieve higher accuracy \citep{panek_guide_2026}. In both cases, the performance further depends on how much the query conditions differ from those during reference mapping \citep{sattler_benchmarking_2018}. Yet scalability and accuracy are both needed in practice: to complement GNSS effectively and bridge GNSS-denied areas, visual localization must scale to entire cities or regions without sacrificing accuracy.

Visual localization methods in outdoor scenarios are generally considered accurate if they achieve translation errors less than 25 cm and rotation errors less than  $2^\circ$ \citep{sattler_benchmarking_2018}. This level of accuracy is sufficient for autonomous driving or AR applications \citep{sarlin_coarse_2019}. However, to complement survey-grade GNSS positioning, which achieves accuracies of a few centimeters (approximately 5 cm in 3D) under favorable conditions, visual localization should yield comparable results. This requirement applies at two levels: firstly, the pose relative to the scene representation, and secondly, the georeferencing of the scene representation itself, which bounds the achievable absolute accuracy. 

To the best of our knowledge, there exists no study that systematically evaluates the accuracy potential of visual localization in street scenes with respect to survey-grade accuracy demands. This is a direct consequence of the lack of publicly available large-scale datasets that contain sufficiently accurate ground-truth poses. In this work, our objective is to address both research gaps. We investigate the following question: what pose accuracy relative to precisely georeferenced street-level reference imagery can sparse feature-based visual localization achieve under favorable conditions? We refer to this as the accuracy potential of visual localization and address it through four sub-questions: (i) how does the selection of reference candidates, guided by pose priors, affect the pose accuracy; (ii) how do image resolution and the choice of local features affect it; (iii) how does it depend on the query camera and on the epoch of the reference imagery; and (iv) how much accuracy is lost by using 2D image-based visual localization compared to a 3D geometry-based method?

Answering these questions requires ground-truth poses whose uncertainty is well below the expected localization errors, which no existing large-scale outdoor dataset provides (Section~\ref{related_work_datasets}). We therefore created a dataset in which reference imagery from two mobile mapping campaigns and query sequences from four different cameras are co-registered with quantified uncertainty. On this dataset, we evaluate the accuracy potential of visual localization using a modular visual localization pipeline built from established components. It uses the precisely georeferenced high-resolution imagery directly as scene representation, selects reference candidates guided by pose priors, and estimates the query pose from a local Structure-from-Motion (SfM) model reconstructed on the fly. Reference imagery acquired by professional vision-based mobile mapping systems is what we refer to as high-end street-level imagery. Our main contributions are summarized as follows:
\begin{itemize}
  \item The FHNW Muttenz dataset\footnote{\url{https://fhnw-muttenz-vl-dataset.github.io/}}: high-resolution reference imagery of a contiguous 10\,km street network from two mobile mapping campaigns about 1.5 years apart, and query sequences from four cameras in five scenes. The two campaigns are co-registered with a 3D RMSE of 5~mm at independent tie points. Co-registration of the query images to the reference imagery yields ground-truth poses accurate to 6--9\,mm in position and well below 0.1$^{\circ}$ in orientation. A sensitivity analysis confirms that these poses are determined by the image observations rather than by the ground control point configuration.
  \item A systematic evaluation of the accuracy potential of sparse feature-based visual localization with respect to reference candidate selection, image resolution, local features, query camera and reference epoch, including a comparison with a representative 3D geometry-based visual localization method. Evaluations at error thresholds down to 1\,cm and 0.1$^{\circ}$ reveal differences between local features that are not resolved at the coarse thresholds of existing benchmarks.
\end{itemize}

The remainder of this article is organized as follows. Section~\ref{related_work} presents related work on visual localization methods and on the datasets used to evaluate them. Section~\ref{methods} introduces the dataset and the visual localization pipeline. Section~\ref{experiments} reports the experimental setup and the results. Section~\ref{discussion} discusses the results and the limitations of this study and identifies directions for future research. Section~\ref{conclusion} concludes the article.

\section{Related work}\label{related_work}

\subsection{Visual localization}\label{related_work_visual_localization}
As outlined in the introduction, visual localization methods can be categorized into \textit{2D image-based} and \textit{3D geometry-based} methods according to their scene representation. The latter are commonly referred to as 3D structure-based \citep{sattler_benchmarking_2018}, which implies that the 3D geometry is derived from SfM. Since many current 3D scene representations are not derived from SfM, we use the term 3D geometry-based instead. As image retrieval (IR) is crucial for both categories, we begin with a brief summary of this area.

\subsubsection{Image retrieval}
IR retrieves images from a database that depict the same scene as a given query image, either to select reference candidates or to approximate the query pose. The image content is typically encoded as a global image descriptor, and similar images are retrieved via a nearest neighbor search in the descriptor space. As shown by \citet{humenberger_investigating_2022}, the appropriate image description method strongly depends on the subsequent pose estimation technique. IR methods originate from both the landmark retrieval (LR) and visual place recognition (VPR) domains, but they are typically tailored to specific environments and tasks and therefore lack generalization \citep{keetha_anyloc_2024, berton_megaloc_2025}.

Early methods based on global image statistics were replaced by methods that aggregate local features \citep{jegou_aggregating_2010}. NetVLAD \citep{arandjelovic_netvlad_2016} established a first baseline for deep global image descriptors, and subsequent work improved on it through novel losses, large-scale datasets and high-level CNN features while yielding smaller descriptors \citep{yu_sp_netvlad_2019, ali-bey_gsv-cities_2022, ali-bey_mixvpr_2023}. CosPlace adopted a training strategy inspired by face recognition and achieved state-of-the-art results \citep{berton_rethinking_2022}. More recently, vision foundation models such as DINOv2 \citep{oquab_dinov2_2024} have been used to design universal solutions \citep{keetha_anyloc_2024, izquierdo_optimal_2024}, at the cost of considerably longer descriptors. Most trained VPR models, however, learn from positives that share the viewpoint of the query image and therefore often fail to retrieve reference images that depict the same scene from different viewpoints.

EigenPlaces \citep{berton_eigenplaces_2023} addresses this limitation with a viewpoint-aware data sampling technique, while also producing compact descriptors. MegaLoc \citep{berton_megaloc_2025} combines this sampling with training on several task-specific datasets and outperforms state-of-the-art VPR and LR models in retrieving reference images with diverse viewpoints. In large-scale environments, retrieval can additionally be constrained by pose priors from mobile sensors, such as GNSS positions and headings \citep{yan_long_term_2023}.

\subsubsection{2D image-based visual localization}
Visual localization methods relying on this representation can be further subdivided into pose approximation, relative pose estimation and regression, and local SfM model construction. In pose approximation, the pose of the query image is adopted from the most similar reference image or interpolated from the poses of the top-\textit{k} retrieved images \citep{torii_247_2015, sattler_understanding_2019}. Relative pose estimation first establishes 2D--2D correspondences between the query and the retrieved images and estimates their relative geometry. Since the reference poses are known, the absolute pose of the query image can then be derived \citep{zhou_learn_2020, dong_lazy_2023}. Relative pose regression, in contrast, directly regresses the pose between two input images \citep{laskar_camera_2017, zhou_learn_2020}. Recent feed-forward 3D reconstruction models \citep{wang_dust3r_2024, leroy_mast3r_2025, wang_vggt_2025} show competitive performance in image matching \citep{image-matching-challenge-2025} and visual localization, and can estimate the query pose even with little visual overlap, where other methods fail \citep{panek_guide_2026}.

Building a local SfM model on the fly from the top-\textit{k} retrieved images and estimating the query pose from 2D--3D matches \citep{sattler_benchmarking_2018, torii_are_2021, humenberger_investigating_2022} combines the properties of 2D image-based and 3D geometry-based methods \citep{panek_guide_2026}. According to \citet{panek_guide_2026}, such approaches achieve the highest pose accuracies among 2D image-based methods owing to their stronger geometric reasoning, but remain slightly less accurate than 3D geometry-based methods. In addition, triangulating 3D points on the fly causes considerable computational overhead.

ImLoc \citep{jiang_imloc_2026} is, to our knowledge, the first 2D image-based approach that surpasses state-of-the-art methods, including those that explicitly exploit 3D scene geometry. It augments the reference images with depth maps and lifts dense RoMa correspondences \citep{edstedt_roma_2024} to 2D--3D correspondences, thereby enabling geometric reasoning without on the fly reconstruction while keeping the scene representation simple. Dense matching, however, is computationally demanding at high image resolutions (Section~\ref{m_image_matching}).

\subsubsection{3D geometry-based visual localization}
Traditionally, scenes for 3D geometry-based visual localization are represented by SfM point clouds \citep{arth_wide_2009, irschara_location_2009, sattler_benchmarking_2018}, and the camera pose is estimated from 2D--3D correspondences obtained by matching query features against the descriptors of the 3D points. To cope with scalability and descriptor ambiguity in large scenes, hierarchical approaches confine matching to the 3D points visible in the top-\textit{k} retrieved reference images \citep{sattler_aachen_2012, sarlin_coarse_2019}. Besides SfM point clouds, meshes \citep{avidan_meshloc_2022, panek_visual_2023, vultaggio_investigating_2024}, 3D city models \citep{majdik_micro-air_2014, loeper_visual_2024}, 3D Gaussian Splatting \citep{leonardis_6dgs_2025, zhai_splatloc_2025} and Neural Radiance Fields (NeRFs) \citep{liu_nerf-loc_2023, zhou_nerfect-match_2024} serve as scene representations, while the pose is still estimated from 2D--3D correspondences derived from feature matching.

Scene coordinate regression methods instead train a network to directly regress the 3D coordinates of image patches, which are then used for RANSAC-based PnP pose estimation \citep{shotton_scene_2013, brachmann_learning_2018, revaud_sacreg_2024}. Absolute pose regression methods predict the query pose directly from the image \citep{kendall_posenet_2015, shavit_abs-pose_2021}, but most of them are not substantially more accurate than pose approximation and generalize poorly to unseen views \citep{sattler_understanding_2019}. Map-relative variants improve generalization to new scenes \citep{chen_map-relative_2024}. Render-and-compare methods iteratively refine a coarse initial pose, for example from IR, by aligning the query image with renderings of the scene representation \citep{chen_inerf_2021, sarlin_back-to-feature_2021, chen_neural-refinement_2024}, but remain inferior to hierarchical feature-based methods in accuracy and robustness \citep{panek_guide_2026}.

All these methods exploit the 3D geometry of the scene, represented explicitly or implicitly, in a globally consistent manner. Precomputing this geometry is time-consuming and resource-intensive, and ensuring global consistency is particularly challenging in large-scale environments \citep{sattler_benchmarking_2018}. Once built, such representations are static and costly to update when the scene changes. Moreover, meshes, city models and learned representations abstract geometric and visual information, which degrades the accuracy of the estimated pose \citep{panek_visual_2023, bieringer_analyzing_2024, gaisbauer_glue_2025, sattler_understanding_2019}.

\subsection{Datasets}\label{related_work_datasets}

\begin{table*}[ht]
\centering

\footnotesize
\caption{Comparison of visual localization datasets: environment (Env.), number of locations (\# Loc.), spatial scale, challenging conditions, camera models, high-resolution imagery, and accuracy of the image poses. Challenges: weather (W), season (S), strong viewpoint changes (V), day/night (N), permanent or temporary construction (C), dynamics such as people, vehicles or rearranged furniture (D), and camera intrinsics (I). Camera models: perspective (P), fisheye (F), spherical (360). Pose accuracy reflects relative accuracy to the reference data, as reported in the original publications.}\label{table_dataset_comparison}
\resizebox{\textwidth}{!}{
\begin{tabular}{m{4.5cm}ccc*{5}{m{0.1cm}}*{2}{m{0.1cm}}ccl}
\toprule
  \multirow{2}{*}{Dataset} & \multirow{2}{*}{Env.} & \multirow{2}{*}{\# Loc.} &  \multirow{2}{*}{Scale} &\multicolumn{7}{c}{Challenging conditions} & Camera model & High res. &Pose\\
  &&&&W&S&V&N&C&D&I& ref / query & images & acc. \\ 
  \midrule
  7-Scenes \citep{shotton_scene_2013} & I & 7& S &&&&&&&& P / P &  &$\approx$ cm \\
  Baidu mall \citep{sun_dataset_2017} & I & 1 & M &&&&&&\checkmark&\checkmark& P / P &  & $\approx$ dm \\
  InLoc \citep{taira_inloc_2021} & I & 5 & M & &&&&&\checkmark&&P / P & \checkmark& $>$ dm \\
  NAVER Labs \citep{lee_large-scale_2021} & I & 5 & M &&&&&&\checkmark&\checkmark&P / P & \checkmark&   $\approx$ dm\\
  NCLT \citep{carlevaris-bianco_university_2016} & I+O & 1& M &\checkmark & \checkmark &&&\checkmark&\checkmark&&P / P& & $\approx$ dm \\
  ETH3D \citep{schops_multi-view_2017} & I+O & 25 & S &&&&&&&& P / P & \checkmark & $\approx$ mm\\
  LaMAR \citep{lamar_dataset_2022} & I+O & 3 & M & \checkmark & & \checkmark & \checkmark & \checkmark & \checkmark & \checkmark &P / P &    & $\approx$ cm \\
  360Loc \citep{huang_360loc_2024} & I+O& 4&  M& \checkmark&&& \checkmark& \checkmark& \checkmark& \checkmark& 360 / P+F+360 & \checkmark &$\approx$ cm  \\
  CroCoDL \citep{blum_crocodl_2025} & I+O & 10 & L & \checkmark && \checkmark& \checkmark& \checkmark& \checkmark& \checkmark& P / P &   &  $\approx$ dm \\ 
  Oxford Day \& Night \citep{wang_seeing_2026} & I+O & 5 & L &&&\checkmark&\checkmark&&\checkmark&&P / P&   & $<$ dm  \\
  San Francisco \citep{chen_city-scale_2011} & O & 1 & L &&&&&\checkmark& \checkmark & \checkmark & P / P &   & $\approx$ m \\
  CMU Seasons \citep{badino_visual_2011, sattler_benchmarking_2018} & O & 17 &  L & \checkmark & \checkmark &&&&\checkmark& \checkmark & P / P &  & $>$ dm\\
  Aachen Day Night \citep{sattler_aachen_2012, sattler_benchmarking_2018} & O & 1 & L & \checkmark &&&& &\checkmark& \checkmark & P / P &   & $>$ dm\\
  Cambridge \citep{kendall_posenet_2015} & O &3 & S &\checkmark &&&&&\checkmark&& P / P &   & $>$ dm \\
  RobotCar Seasons \citep{maddern_robotcar_2017, sattler_benchmarking_2018} & O & 49&  L & \checkmark & \checkmark & & \checkmark & &\checkmark &  & P / P &   & $\approx$ dm\\
  CrowdDriven \citep{jafarzadeh_crowddriven_2021} & O & 26& M & \checkmark& \checkmark& \checkmark& \checkmark& \checkmark& \checkmark& \checkmark & P / P &   & $<$ dm\\
  SenseLoc \citep{yan_long_term_2023} & O & 1& M & \checkmark & \checkmark & & \checkmark& \checkmark& \checkmark& &P / P &   & $<$ dm \\
  \midrule
  \textbf{FHNW Muttenz (Ours)} & O& 1& L & & \checkmark &\checkmark&& \checkmark & \checkmark & \checkmark & P / P+F & \checkmark   &  $<$ cm  \\
\bottomrule
\end{tabular}}
\end{table*}

Visual localization datasets are typically collected with specific applications and tasks in mind and therefore vary in scene type, sensor configuration and intended use. Table~\ref{table_dataset_comparison} provides an overview of popular datasets and benchmarks currently used in the field. Challenging conditions, which can severely degrade the performance of visual localization methods, occur more frequently outdoors than indoors, where they are largely limited to dynamics and changing camera intrinsics.

Reliable evaluation requires precise ground-truth poses, whose estimation is itself limited by challenging conditions and by the spatial extent of the scene. For small- to medium-scale environments, a common strategy is to register the reference and query sequences against high-precision LiDAR data (ETH3D \citep{schops_multi-view_2017}, NAVER Labs \citep{lee_large-scale_2021}, LaMAR \citep{lamar_dataset_2022}, CroCoDL \citep{blum_crocodl_2025}). Matches to the scanner's panoramic images or to rendered views of the colorized point cloud are lifted to 2D--3D correspondences through the associated 3D points, which then drive pose estimation and refinement \citep{schops_multi-view_2017, taira_inloc_2021, lamar_dataset_2022}. Alternatively, multi-sensor platforms are combined with SLAM algorithms, whose maps provide an initial alignment that is refined in post-processing \citep{yan_long_term_2023, huang_360loc_2024}. Large-scale outdoor datasets such as San Francisco \citep{chen_city-scale_2011}, CMU Seasons \citep{badino_visual_2011, sattler_benchmarking_2018} and RobotCar \citep{maddern_robotcar_2017, sattler_benchmarking_2018} are typically acquired with mobile mapping systems (MMS) \citep{puente_review_2013}. Their ground-truth poses are usually obtained by orienting individual sequences via SfM and co-registering them via ICP or manually measured tie points, incorporating relative sensor orientations and approximate poses from direct sensor orientation where available \citep{chen_city-scale_2011, sattler_benchmarking_2018}.

Although query images are often captured with cameras that differ from those used for reference mapping, cross-device localization has long been neglected. Most datasets reproject images from non-perspective cameras to the classical pinhole model \citep{carlevaris-bianco_university_2016, maddern_robotcar_2017, wang_seeing_2026}. 360Loc \citep{huang_360loc_2024} provides query images following the perspective, fisheye and spherical camera models, but its pinhole and fisheye images are generated with virtual cameras rather than recorded by physical devices.

High-resolution imagery is also underrepresented. Since computational resources and latency are often crucial in visual localization applications, only ETH3D \citep{schops_multi-view_2017}, InLoc \citep{taira_inloc_2021}, NAVER Labs \citep{lee_large-scale_2021} and 360Loc \citep{huang_360loc_2024} provide high-resolution images, none of which covers a large-scale outdoor environment. Moreover, most datasets consist of several non-overlapping submaps. While this design is reasonable when aggregating distinct environments and estimating ground-truth poses \citep{sattler_benchmarking_2018}, real-world environments cannot be partitioned in this way without introducing edge cases at the boundaries.

Long-term localization performance is likely to decrease as the scene gradually changes \citep{sattler_benchmarking_2018, zhang_visual_2026}, which makes updating the reference map critical, regardless of the scene representation. However, available datasets usually provide a single reference sequence and therefore do not support investigations on how to update reference maps efficiently.

Further datasets address specific scene representations. The cadloc dataset \citep{panek_visual_2023} and egenioussBench \citep{fanta-jende_egenioussbench_2025} provide benchmarks for mesh-based visual localization, relying on 3D models from the internet that represent reality only imperfectly and on geospatial meshes, respectively. The TUM2TWIN dataset \citep{wysocki_tum2twin_2026} provides multi-sensor data for universal digital twins, co-registered with an accuracy in the upper centimeter to decimeter range, and is therefore suitable for investigations on scene representations such as city models or meshes.

In summary, a diverse set of datasets is available for developing and assessing visual localization methods. However, none of them combines a large-scale street-scene setting with ground-truth poses accurate enough to assess the accuracy potential of visual localization.

\section{Materials and methods}\label{methods}
\subsection{Dataset}\label{method_dataset}
The dataset is designed to evaluate the accuracy potential of visual localization in large-scale outdoor environments and addresses the limitations identified in Section~\ref{related_work_datasets}.

\subsubsection{Study area}

\begin{figure*}[ht]
\includegraphics[width=\textwidth]{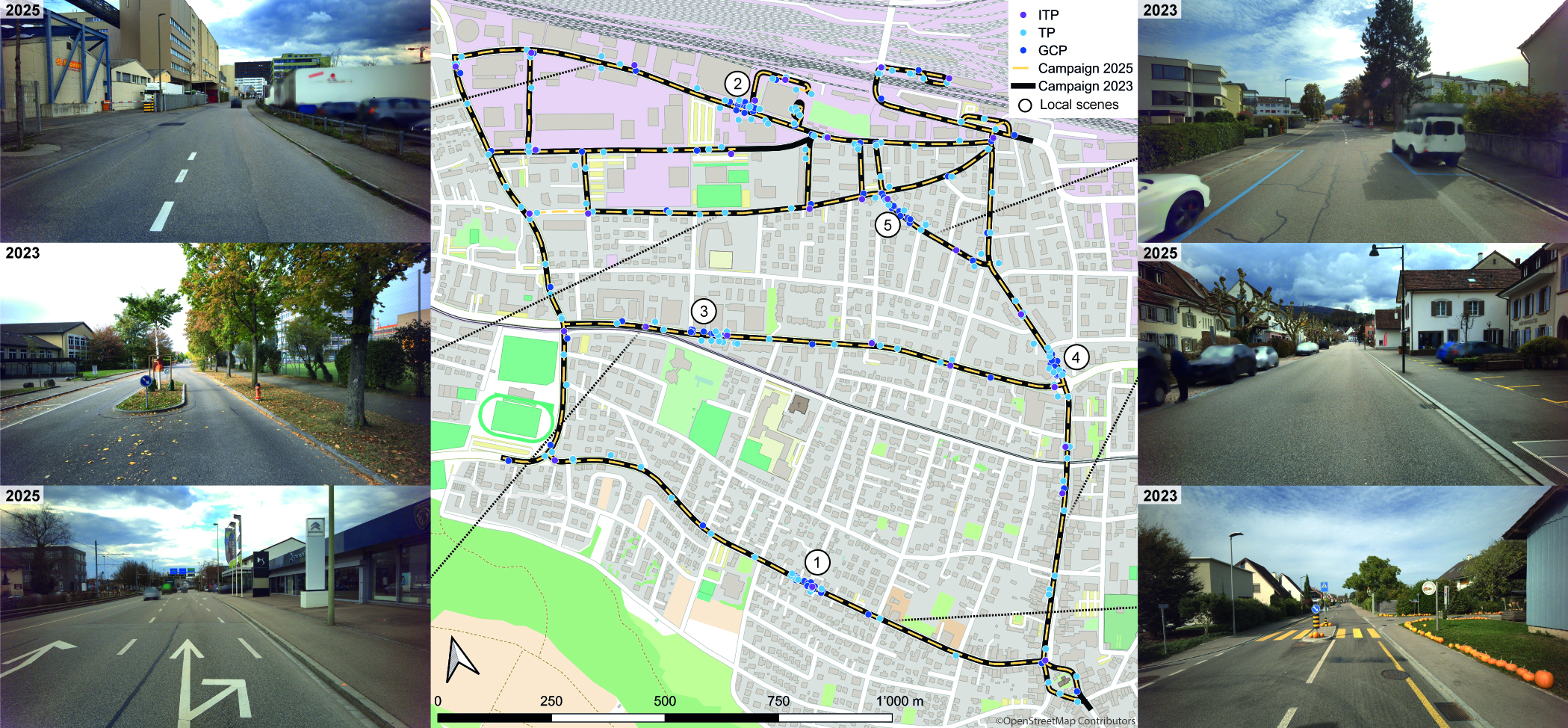}
\caption{Study area in Muttenz, Switzerland. Street network of the two mobile mapping campaigns, GCPs, TPs and ITPs used for georeferencing and co-registration, and the five query scenes (circled numbers). Example images illustrate the conditions on site. Background map: \copyright\ OpenStreetMap contributors.}\label{fig_study_area}
\end{figure*}

The study area, depicted in Figure~\ref{fig_study_area}, encompasses various districts of Muttenz, a suburban municipality near Basel, Switzerland. The areas covered include parts of a historic town center, commercial areas, residential areas with detached houses, multi-story apartment buildings, diverse vegetation, and a mixed industrial area with production facilities and brownfield sites with parked vehicles and trucks. The road network, with a length of approximately 10 km, exhibits substantial heterogeneity, encompassing multi-lane roads, intersections with tram lines, wide pedestrian sidewalks, and ordinary two- and single-lane roads in residential areas (Figure~\ref{fig_study_area}).

\subsubsection{Reference imagery}\label{m_dataset_reference_imgs}
To efficiently acquire reference imagery along the street network, we employ vision-based mobile mapping. The entire study area was mapped twice. The first campaign was conducted in autumn (October) 2023. The weather conditions during the first campaign were predominantly sunny, resulting in shadow patterns, and the trees showed slightly tinted foliage. The second campaign was conducted in early spring (March) 2025, under overcast skies, resulting in diffuse light, with trees free of foliage (Figure~\ref{fig_study_area}). During both campaigns, extensive construction work was in progress, including the complete redesign of entire road sections.

\begin{table}[ht]
\centering
\caption{Camera systems of the 2023 and 2025 mobile mapping campaigns with horizontal (h) and vertical (v) field of view (FoV),  resolution and number of images. Note: for the panorama camera the single camera heads are reported.}\label{mms_sensor_setup}
\setlength{\tabcolsep}{4pt}
\begin{tabular}{llcccr}
\toprule
& \multirow{2}{*}{Camera system} & \multicolumn{2}{c}{FoV [$^\circ$]}  & \multirow{2}{*}{Resolution} & \# \\
& & h & v &&  Images \\ \midrule
\parbox[t]{2mm}{\multirow{3}{*}{\rotatebox[origin=c]{90}{2023}}} & Stereo front & 96.0 & 64.1 & 5472 $\times$ 3084 & 8054 \\
& Stereo back left  & 86.9 & 64.4 & 4864 $\times$ 3232 & 8054 \\ 
& Stereo back right & 86.9 & 64.4 & 4864 $\times$ 3232 & 8054 \\ 
& Pano camera (5$\times$) & 73.0 & 86.0 & 1536 $\times$ 1936 & 4027\\ \midrule
\parbox[t]{2mm}{\multirow{2}{*}{\rotatebox[origin=c]{90}{2025}}} & Mono front & 75.1 & 46.9 & 5472 $\times$ 3084 & 4170 \\ 
& Pano camera (5$\times$) & 73.0 & 86.0 & 2918 $\times$ 3677 & 4170 \\
\bottomrule
\end{tabular}%
\end{table}

\begin{figure}[ht]
\includegraphics[width=\linewidth,keepaspectratio]{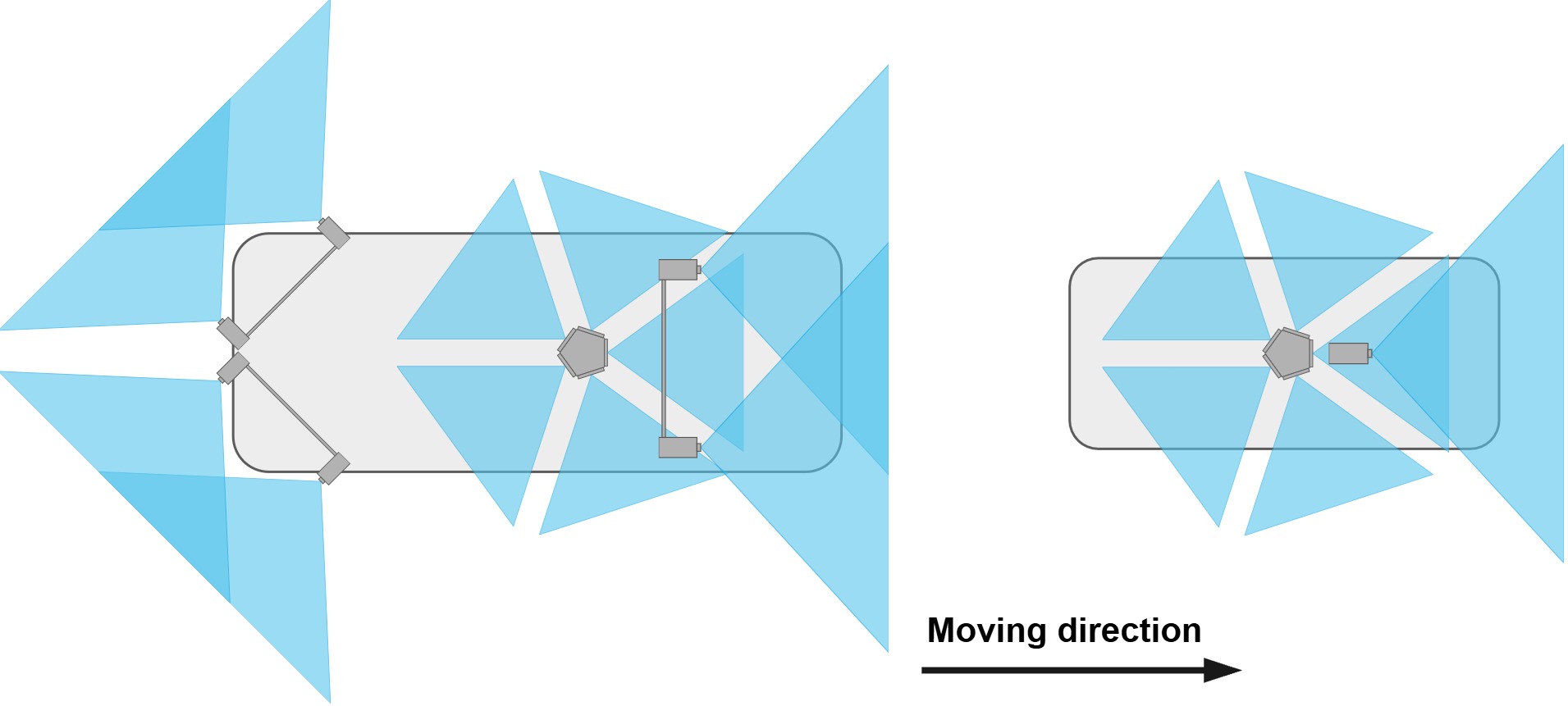}
\caption{Mobile mapping systems of the 2023 (left) and 2025 (right) campaigns, drawn to scale.}\label{mms_cam_config}
\end{figure}

In the 2023 campaign, the camera setup (Table~\ref{mms_sensor_setup} \& Figure~\ref{mms_cam_config}, left) consisted of three high-resolution stereo systems facing the front, back right, and back left, combined with a medium-resolution multi-head panorama camera, Ladybug5+. In 2025, the MMS was equipped with a high-resolution monocular camera facing the front and a high-resolution multi-head panorama camera, Ladybug6 (Table~\ref{mms_sensor_setup} \& Figure~\ref{mms_cam_config}, right). Figure~\ref{mms_cam_config} visualizes the camera footprints for each campaign. From 2023 to 2025, both the camera sensor setup and the size of the mapping platform changed. As a result, rear-view images from the 2025 panoramic camera contain almost no elements of the MMS itself, unlike the 2023 images. During the reference mapping campaigns, all cameras were triggered synchronously at approximately 4 m intervals. Wide roads were traversed in both directions, while narrow roads were traversed only once. Construction activities slightly changed the road network coverage between campaigns (Figure~\ref{fig_study_area}).

All camera systems used were pre-calibrated, yielding distortion-free images that were, in the case of the stereo systems, also rectified. The panoramic images are only available as cube maps with stitching errors caused by the differing projection centers of the multi-head panorama cameras. Hence, we remapped them to the original camera heads, resulting in five images with a horizontal field of view (FoV) of 73$^\circ$ (Table~\ref{mms_sensor_setup}), with the image content covering 72$^\circ$ per head, so that the five images cover the full panorama without overlap.

For georeferencing of the acquired imagery, both MMS were equipped with a tactical-grade IMU and a geodetic GNSS receiver comparable to the MMS described by \citet{cavegn_robust_2018}. The image poses were computed via integrated georeferencing \citep{eugster_integrated_2012}, using 32 of the ground control points (GCPs) described below, evenly distributed throughout the mapping perimeter, which yields a georeferencing accuracy in the sub-decimeter range. Evaluating visual localization at centimeter thresholds requires sub-centimeter image poses, obtained in two further steps. 

First, we applied image-based georeferencing \citep{cavegn_robust_2018} to each campaign independently to obtain a geometrically stable photogrammetric model and compensate for residual inaccuracies in the image poses. Using Agisoft Metashape \citep{agisoft_metashape_2024}, we refined all image poses within a bundle adjustment. Stereo systems and panorama cameras were modeled as multi-camera systems using relative orientation constraints. Each camera was assigned its initial pose from integrated georeferencing to efficiently select image pairs for image matching.

\begin{table}[ht]
\centering
\caption{Co-registration of the two mobile mapping campaigns: RMSE of the coordinate differences between 2023 and 2025 at the GCPs measured at both campaigns, TPs and ITPs.}\label{corregistration_results_mm_campaigns_rel}
\begin{tabular}{lS[table-format=3.0]*{4}{S[table-format=1.4]}}
\toprule
Point&Num& \multicolumn{4}{c}{RMSE [m]} \\
type& points&{X}&{Y}&{Z}&{3D}\\ \midrule
GCP&43&0.001&0.002&0.001&0.002\\
TP&180&0.004&0.004&0.002&0.006\\
ITP&29&0.003&0.003&0.002&0.005\\ \bottomrule
\end{tabular}
\end{table}

Second, for co-registration of both mobile mapping campaigns, 69 GCPs determined using RTK-GNSS and 180 clearly identifiable tie points (TPs) (Figure~\ref{fig_study_area}, blue points) were manually measured in approximately 10 images per campaign. Of the 69 GCPs, 66 were measurable in the 2023 and 46 in the 2025 imagery, 43 of them in both campaigns. A global bundle adjustment was then performed, incorporating all GCPs, the TPs and the initial 6-DoF image poses from integrated georeferencing. No automatic correspondences between images of different campaigns entered this adjustment, so that the two campaigns are connected exclusively via the manually measured GCPs and TPs. For the GCPs the 3D positional standard deviation was set to 5 cm, while the image poses were incorporated with a 3D position standard deviation of 10 cm and a 3D rotation standard deviation of 1$^\circ$.

To validate the accuracy of co-registration, an additional set of 29 independent tie points (ITPs) (Figure~\ref{fig_study_area}, purple points) uniformly distributed throughout the mapping perimeter was measured in the images of both campaigns. These ITPs were excluded from the bundle adjustment to ensure independence. The 3D coordinates of the GCPs, TPs, and ITPs were then triangulated independently for each mobile mapping campaign from the fixed image poses. For each point set, we report the root mean square error (RMSE) of the coordinate differences between the 2023 and 2025 campaigns (Table~\ref{corregistration_results_mm_campaigns_rel}). The two campaigns are co-registered in the sub-centimeter range, with the ITPs providing an independent check. The absolute georeferencing accuracy of each campaign, determined from the residuals of the GCPs, amounts to a 3D RMSE of 0.025\,m for both the 2023 (66 GCPs) and the 2025 campaign (46 GCPs). Since these points were used in the bundle adjustment, this is an internal measure of accuracy.

\subsubsection{Query sequences}\label{m_dataset_query_sequences}

\begin{figure*}[ht]
\includegraphics[width=\textwidth]{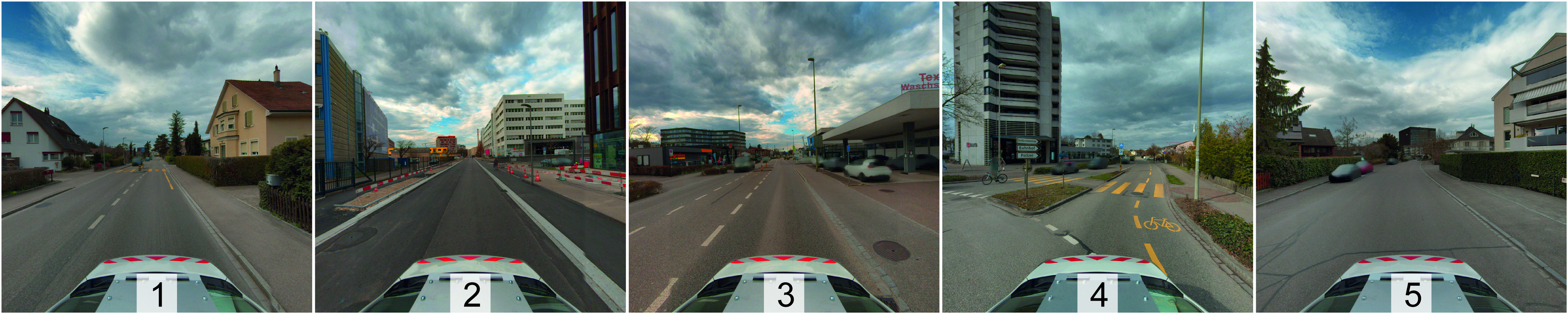}
\caption{Example images of the five query scenes, which differ in street type, environment and scene changes such as moving cars, construction sites and vegetation.}\label{fig_study_area_scenes}
\end{figure*}

For query image acquisition, we identified five small-scale scenes throughout the study area (Figure~\ref{fig_study_area}, circled numbers). Each of these scenes has different characteristics (Figure~\ref{fig_study_area_scenes}). Scene 1 depicts a local access road in a suburban environment, with sidewalks on both sides of the street and a pedestrian crossing with a small island. The street traverses a residential neighborhood of detached houses and abundant greenery, including hedges and trees. In scene 2, a local access road in an area characterized by multi-story commercial, industrial, and office buildings is shown. Road construction is ongoing on the street and in its immediate surroundings; as a result, no road markings are visible and numerous temporary construction fences are in place. On both sides of the road, there are wide sidewalks and some unfinished flower beds without plants. Scene 3 represents a busy, multi-lane main road in a commercial environment, featuring car dealers and a fuel station with predominantly glass and other reflective façades, and many dynamic objects, such as cars. On both sides of the street, there is a sidewalk and only a small amount of vegetation. In scene~4, a local access road with an integrated cycle path traverses a neighborhood with multi-story urban residential and commercial buildings. Sidewalks are provided on both sides of the street, separated by flower beds. The environment is densely vegetated, with trees and hedges. Scene 5 shows a narrow residential street in a neighborhood of detached houses, with plenty of greenery, including large trees and hedges. Various cars are parked on the side of the road.

As query sensors, we employed four camera configurations with different characteristics, including resolution, field of view, and camera models. The four camera systems are: a Canon EOS R with 24 mm and 35 mm lenses, a GoPro Hero 7, and an iPhone 14 Pro (Table~\ref{sensors_query_sequences}).

\begin{table}[ht]
\centering
\caption{Query cameras with horizontal (h) and vertical (v) field of view (FoV), projection model (P: perspective, F: fisheye) and sensor resolution.}\label{sensors_query_sequences}
\begin{tabular}{l*{2}{S[table-format=3.1, table-number-alignment=center]}cc}
\toprule
\multirow{2}{*}{Camera} & \multicolumn{2}{c}{FoV [$^\circ$]} & {Cam} & \multirow{2}{*}{Resolution} \\
 & {h} & {v} & {model} & \\ \midrule
Canon 24mm& 72.3& 51.9 & P & 6720 $\times$ 4480 \\
Canon 35mm& 53.3& 37.0 & P & 6720 $\times$ 4480 \\ 
GoPro Hero 7 & 122.6& 94.4 & F & 4000 $\times$ 3000 \\ 
iPhone 14 Pro & 57.1&71.9 & P & 1440 $\times$ 1920 \\
\bottomrule
\end{tabular}
\end{table}

The query image acquisition was conducted in late winter (February) 2025. The weather conditions varied. Two scenes were captured under predominantly cloudy conditions, while the remaining three scenes were captured in sunny weather. At that time, the trees were free of foliage. In each scene, a sequence of images was captured with all four cameras. The sequences differ in trigger rate, trajectory, length, and number of images because each was captured one after another. The Canon camera was triggered manually, ensuring the images were captured in roughly an upright orientation. For the GoPro, an upright configuration was used as well. Instead of manual triggering, a video was recorded at 24 frames per second (FPS) with electronic stabilization disabled. For iPhone 14 Pro sequences, PIX4DCatch \citep{pix4dcatch} was used and image triggering was set according to distance and angle criteria of 0.1 m and 15$^\circ$, respectively. During this capture, the iPhone was deliberately pointed slightly towards the ground and, at times, moved rapidly to create a trajectory similar to walking while using an AR navigation application. Each sequence forms one or more closed loops. The GoPro and iPhone sequences were subsampled to every 10th and every 5th image, respectively, to obtain image counts comparable to the Canon sequences. The number of query images per camera and scene is listed in Table~\ref{tbl_method_dataset_n_query_images}. Furthermore, for each scene, 7–11 clearly identifiable and distinct natural points were selected as control points (CPs) and determined using RTK-GNSS.

\begin{table}[ht]
\centering
\caption{Number of query images per camera and scene.}\label{tbl_method_dataset_n_query_images}
\begin{tabular}{lrrrrrr}
\toprule
\multirow{2}{*}{Camera} & \multicolumn{5}{c}{Scenes} & \multirow{2}{*}{Total} \\
& 1 & 2& 3 & 4 & 5& \\ \midrule
Canon 24mm & 165&194&135&142&165&801 \\
Canon 35mm & 154&158&116&168&181&777 \\ 
GoPro Hero 7 & 415&376&370&383&403&1947 \\ 
iPhone 14 Pro & 435&246&350&380&463&1874\\ \midrule
Total & 1169&974&971&1073&1212&\bfseries5399\\ \bottomrule
\end{tabular}
\end{table}

For co-registration of the acquired query images with the reference imagery, we employed a multi-stage SfM-based procedure using Agisoft Metashape \citep{agisoft_metashape_2024}. Rather than matching query against reference images directly, which \citet{zhang_reference_2021} identify as the main difficulty in SfM-based reference pose generation, we first create a stable photogrammetric model of the query sequences and then co-register it to the reference imagery via clearly identifiable and precisely measurable points in object space.

To create the photogrammetric model, images of all four sequences were initially aligned individually. For each, at least three CPs were measured to transform the image poses into a unified global reference frame. We then performed an additional image alignment using the image poses from the first alignment to efficiently select matching candidates across sequences. All CPs, as well as approximately 30 additional distinctive tie points at strategically selected locations, were measured in approximately 15 images from different camera sequences. These points help to mutually stabilize the sequences from different cameras per scene.

To co-register the photogrammetric model with the reference imagery, we manually measured approximately 19--24 clearly identifiable points in both the reference and the query images. Each point was observed from as many different viewing directions as possible and their 3D coordinates were triangulated from the fixed reference image poses. For each scene, at least three points were selected as independent check points and excluded from the subsequent adjustment. Since the co-registration refers to the reference imagery, we term these check points relative check points (RCPs) and the points used for co-registration relative ground control points (RGCPs). 

Finally, we performed a bundle adjustment that jointly optimizes the poses of the query images and their camera calibration parameters, with the RGCPs introduced as control points with a 3D standard deviation of 0.005 m. The CPs served only for the initial alignment and were therefore not introduced in the bundle adjustment. Query and reference images are thus connected exclusively via the manually measured points. The camera model comprises the focal length $f$, the principal point $(c_x, c_y)$, and the radial ($k_1$--$k_4$) and decentering ($p_1$, $p_2$) distortion coefficients of the Brown--Conrady model \citep{brown_calib_1971}, applied to a perspective projection for the Canon and iPhone cameras and to an equidistant fisheye projection for the GoPro \citep{agisoft_metashape_2024}. The affinity and skew terms $b_1$ and $b_2$ of the extended model \citep{fraser_self_calib_1997} were not estimated, assuming square pixels and orthogonal image axes, and because $b_1$ is highly correlated with the focal length.

\begin{table}[ht]
\centering
\caption{Co-registration of the query images per scene: RMSE of the coordinate differences between points triangulated from the query and from the reference image poses. RCPs were excluded from the adjustment and serve as independent check points.}\label{corregistration_results_local_scenes}
\begin{tabular}{lcS[table-format=2.0]*{4}{S[table-format=1.3]}}
\toprule
\multirow{2}{*}{Scenes} & \multicolumn{2}{c}{Points} & \multicolumn{4}{c}{RMSE [m]}\\
& {type}& {\#} & X & Y & Z & {3D}\\ \midrule
\multirow{2}{*}{Scene 1} & RGCP & 16 & 0.005 & 0.006 & 0.002 & 0.008 \\
& RCP & 3 & 0.002 & 0.006 & 0.003 & 0.007\\ 
\multirow{2}{*}{Scene 2} & RGCP & 18 & 0.007 & 0.002 & 0.004 & 0.008 \\
& RCP & 4 & 0.006 & 0.004 & 0.004 & 0.008 \\ 
\multirow{2}{*}{Scene 3} & RGCP & 20 & 0.004 & 0.005 & 0.003 & 0.008 \\
& RCP & 4 & 0.007 & 0.004 & 0.003 & 0.009\\ 
\multirow{2}{*}{Scene 4} & RGCP & 16 & 0.005 & 0.006 & 0.002 & 0.008 \\
& RCP & 4 & 0.003 & 0.003 & 0.004 & 0.006\\
\multirow{2}{*}{Scene 5} & RGCP & 17 & 0.004 & 0.004 & 0.002 & 0.006 \\
& RCP & 6 & 0.004 & 0.004 & 0.002 & 0.006 \\
\bottomrule
\end{tabular}
\end{table}

To assess the co-registration, we triangulated the 3D coordinates of the RGCPs and RCPs from the fixed query image poses and report the RMSE of their differences to the coordinates triangulated from the fixed reference image poses (Table~\ref{corregistration_results_local_scenes}). For the RCPs, which were excluded from the adjustment, these differences provide an independent check. At 6--9\,mm, they show that the photogrammetric models are co-registered to the reference imagery with sub-centimeter accuracy. This is an upper bound, since the differences also contain the uncertainty in manually measuring the natural points in the images.

The differences between the triangulated and the RTK-GNSS coordinates of the CPs provide an independent check of the georeferencing of the reference imagery, because the final adjustment is tied to it exclusively via the RGCPs. They amount to 1.4--3.0\,cm per scene (Table~\ref{georeferencing_results_local_scenes}) and 2.2\,cm over all 48 CPs, consistent with the internal estimate of 2.5\,cm (Section~\ref{m_dataset_reference_imgs}), and likewise represent an upper bound, since they additionally contain the co-registration error and the uncertainty of the RTK-GNSS measurements. For consistent reporting, we use 2.5\,cm as the georeferencing accuracy of the entire dataset in the following.

\begin{table}[ht]
\centering
\caption{Georeferencing accuracy of the query images per scene: 3D RMSE between the CP coordinates from RTK-GNSS and those triangulated from the fixed query image poses.}\label{georeferencing_results_local_scenes}
\begin{tabular}{lccccc}
\toprule
 &  & \multicolumn{4}{c}{Scenes}\\
& {1}& {2} & {3} & {4} & {5}\\ \midrule
Num points & 11 & 10 & 7 & 9 & 11 \\
RMSE [m] & 0.014 & 0.024 & 0.030 & 0.020 & 0.020 \\
\bottomrule
\end{tabular}
\end{table}

\subsubsection{Uncertainty of the query ground-truth poses}\label{m_dataset_query_sensitivity}

Assessing the uncertainty of the individual query poses directly would require reference poses of superior accuracy, which can only be obtained with considerable effort and thus for a few images at most, not for all 5399 query images from four different cameras (Table~\ref{tbl_method_dataset_n_query_images}). The co-registration accuracy in Table~\ref{corregistration_results_local_scenes} does not close this gap, as it characterizes the photogrammetric models as a whole rather than the single image poses. Moreover, camera calibration parameters and image poses are correlated when estimated simultaneously \citep{remondino_digital_2006}. Thus, errors in one can be compensated by the other, biasing the poses while leaving the object-space residuals small. We therefore assess how sensitive the image poses and the camera calibration are to the control point configuration. This configuration defines the datum of the poses and thus the frame in which all pose errors are evaluated. If the photogrammetric model is geometrically well conditioned by the image observations, a change of the RGCP configuration mainly changes the realization of the datum. Therefore, the residual pose differences, i.e., the differences remaining after transforming the poses into a common datum, are small, and the camera calibration remains stable. Large residual differences and varying calibration parameters would instead indicate that the poses are determined by the control point configuration rather than by the image observations.
 
For each scene, we pooled the RGCPs and RCPs and repeated the co-registration described in Section~\ref{m_dataset_query_sequences} 30 times. Each run used randomly drawn subsets of 75\,\%, 50\,\% and 25\,\% of the points as RGCPs, while the remaining points serve as RCPs. To avoid a favorable starting point, the camera calibration was initialized in each run with nominal values: focal length from the manufacturer specifications, principal point at the image center, all other parameters set to zero. The resulting poses were transformed into the datum of the published ground truth, and the residual pose differences were computed separately for position and orientation. In addition, we assessed the variation of the calibration parameters.

\begin{table}[t]
\centering
\caption{Residual pose differences to the published ground truth after removing a common similarity transformation; 30 runs per scene with 75\,\%, 50\,\% and 25\,\% of the points as control. $\Delta t$: distance between the projection centers; $\Delta R$: relative rotation angle (Eq.~\ref{formula_e_r}). p95: per run over all images, averaged over the runs; max: over all images and runs.}
\label{tab_gt_img_pose_sensitivity}
\begin{tabular}{c ccc ccc}
\toprule
& \multicolumn{3}{c}{$\Delta t$ [mm]} & \multicolumn{3}{c}{$\Delta R$ [$10^{-3}\,{}^\circ$]} \\
{Scene} & 75\,\% & 50\,\% & 25\,\% & 75\,\% & 50\,\% & 25\,\% \\
\midrule
\multicolumn{7}{c}{\textit{Quantile 95 \% (p95)}} \\
1 & 0.6 & 1.3 & 1.5 & 0.5 & 0.8 & 1.0 \\
2 & 0.6 & 1.0 & 1.4 & 0.9 & 1.4 & 1.8 \\
3 & 2.5 & 2.2 & 4.8 & 1.7 & 2.1 & 3.5 \\
4 & 0.5 & 0.5 & 0.5 & 0.4 & 0.4 & 0.6 \\
5 & 1.3 & 2.0 & 2.7 & 0.5 & 0.8 & 1.1 \\
\midrule
\multicolumn{7}{c}{\textit{Maximum value (max)}} \\
1 & 2.5 & 3.5 & 3.1 & 1.5 & 2.2 & 2.0 \\
2 & 2.2 & 3.5 & 3.8 & 3.2 & 4.0 & 4.3 \\
3 & 4.6 & 5.0 & 11.2 & 6.5 & 5.8 & 9.4 \\
4 & 1.0 & 1.2 & 1.2 & 1.6 & 1.5 & 1.5 \\
5 & 4.9 & 5.5 & 5.0 & 2.1 & 2.4 & 2.4 \\
\bottomrule
\end{tabular}
\end{table}

The residual pose differences are at the millimeter level (Table~\ref{tab_gt_img_pose_sensitivity}). With 75\,\% and 50\,\% of the points as RGCPs, the p95 values do not exceed 2.5\,mm and 0.0021$^\circ$, and reducing the control from 75\,\% to 50\,\% increases them by at most 0.7\,mm and 0.0005$^\circ$. Even with only 25\,\% of the points, all runs converged and the p95 values do not exceed 4.8\,mm and 0.0035$^\circ$. The largest differences occur in Scene~3 (up to 11.2\,mm and 0.0094$^\circ$ with 25\,\% control), where all sequences were captured from one sidewalk, resulting in the weakest network geometry of all scenes. The camera calibration remains stable as well. Across all runs, the focal length varies by at most 38\,ppm (0.174\,px) and the principal point by at most 55\,ppm (0.370\,px) in x and 57\,ppm (0.257\,px) in y direction.
 
Reducing the number of control points therefore mainly changes the realization of the datum, while the internal geometry of the photogrammetric network and the camera calibration remain largely unaffected. A compensation between calibration and image poses is therefore not evident, and the poses within the photogrammetric model are determined by the image observations rather than by the control point configuration. The resulting uncertainty of the ground-truth positions relative to the reference imagery is 6–9\,mm. This estimate combines the RMSE of the co-registration determined by the RCPs (Table~\ref{corregistration_results_local_scenes}) with the p95 residual pose differences when using  75\,\% of the RGCPs via error propagation. The orientation uncertainty is well below 0.1$^\circ$. 

This analysis characterizes the precision of the ground-truth poses and their sensitivity to the realization of the datum and to the initialization of the camera calibration. Since the image observations, the manually measured points and the processing chain are identical in all runs, it cannot detect errors common to them. The resulting residual uncertainty is discussed in Section~\ref{dis_limitations}.

\subsection{Visual Localization Pipeline}\label{method_vl_pipeline}
\begin{figure*}[ht]
\includegraphics[width=\textwidth,height=\textheight, keepaspectratio]{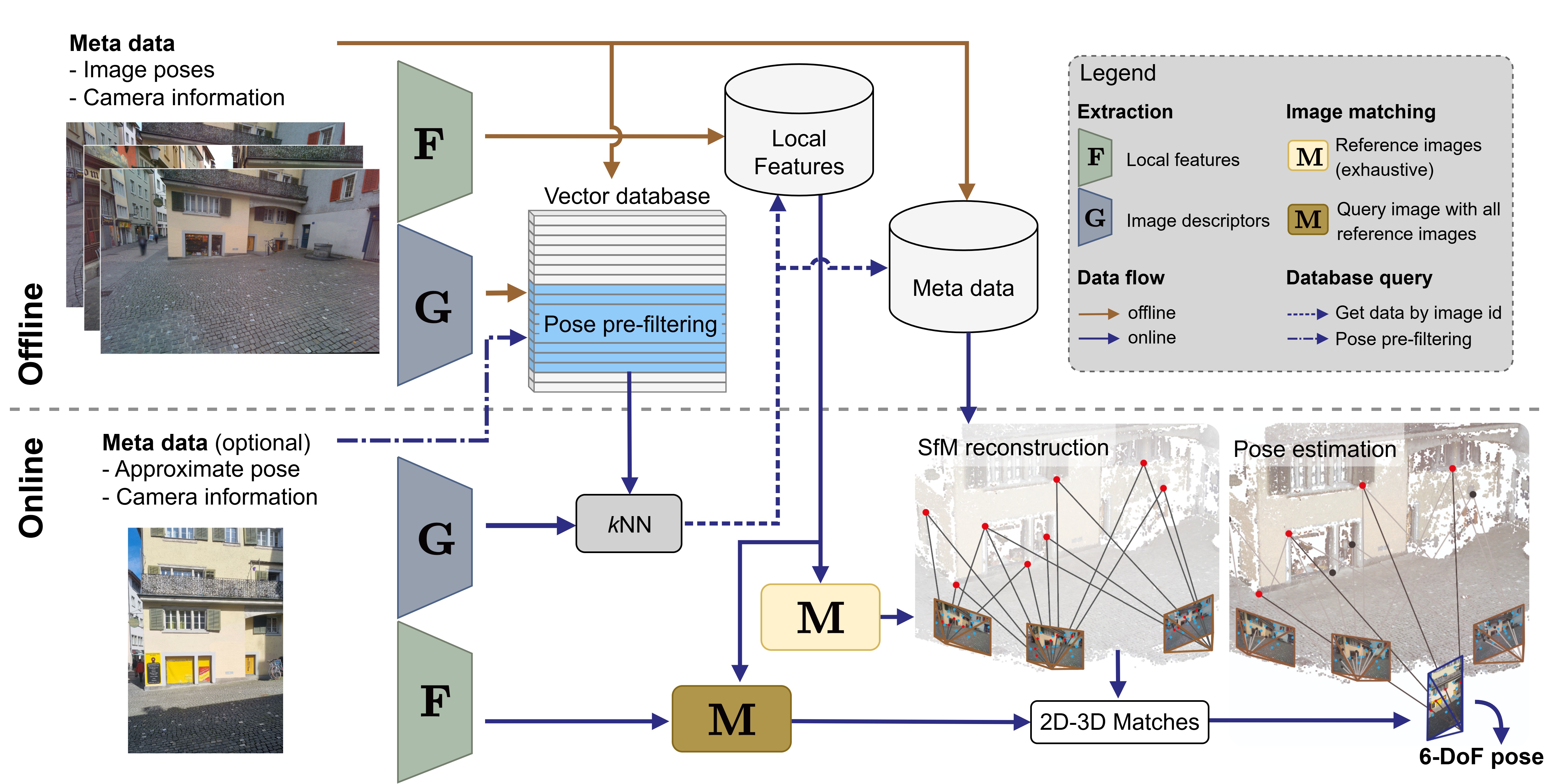}
\caption{Schematic overview of the visual localization pipeline, with offline preprocessing of the reference imagery (upper half) and online localization of a query image (lower half); see Section~\ref{method_vl_pipeline} for details.}\label{fig_vl_pipeline}
\end{figure*}
Figure~\ref{fig_vl_pipeline} provides a schematic overview of our visual localization pipeline. Its central design choice is to use precisely georeferenced, high-resolution imagery acquired by vision-based MMS directly as the scene representation. This choice is motivated by the need for scalability, accuracy, and the flexibility to modify individual components of the visual localization pipeline. The pipeline is composed  of established components, which we adopt with minor adaptations: prior-guided reference candidate selection \citep{yan_long_term_2023}, tiled extraction of sparse local features \citep{maiwald_solving_2023, morelli_deep-image-matching_2024}, and pose estimation from a local SfM model reconstructed on the fly \citep{torii_are_2021, humenberger_investigating_2022}. Sparse local features are chosen because, unlike dense matchers, they remain manageable at high image resolutions and large scale (Section~\ref{m_image_matching}).

The pipeline consists of two stages. In the offline stage, global image descriptors and local features are extracted from the reference images. The global descriptors are stored in a vector database together with the image poses, and the local features and metadata, such as image poses and camera intrinsics, in separate databases (Figure~\ref{fig_vl_pipeline}, upper half). The online stage comprises three steps, which are described in the following subsections: reference candidate selection, feature extraction and matching, and local SfM model construction and pose estimation (Figure~\ref{fig_vl_pipeline}, lower half). The specific methods and their settings are given in Section~\ref{exp_implementation_details}.

\subsubsection{Reference candidate selection}\label{m_ref_cand_selection}
The coarse-to-fine localization paradigm proposed by \citet{sarlin_coarse_2019} is widely adopted in visual localization methods. Although the retrieval of reference candidates has proven highly effective, these methods remain susceptible to failure when images depict different locations while exhibiting nearly identical visual content \citep{cai_doppelgangers_2023}, a problem that grows with scale. To disambiguate the similarity search in large-scale environments and to improve efficiency, we add a pose pre-filtering step to perform IR on a spatially constrained subset of descriptors. Since many devices, such as smartphones, are nowadays equipped with a GNSS sensor, we adopt the sensor-guided retrieval procedure introduced by \citet{yan_long_term_2023} to constrain the search space. Along with each image descriptor, we store position and heading information as additional attributes in the vector database. 
In the online stage, an axis-aligned square search region, extending $\pm d$ around the approximate query position, where d is the positional uncertainty, is used to retrieve the candidate descriptors from the local neighborhood. This subset can be further refined by a heading filter. In the absence of explicit information about pose uncertainty, a default bounding box of 1$\times$1 km ($d$ = 500 m) is adopted for spatial filtering. Finally, within the resulting subset, a k-nearest-neighbor (kNN) search identifies the \textit{k} most similar reference  images. If no position prior is available, spatial filtering is omitted and the k-nearest-neighbor search covers the entire database. Since the local SfM model requires reference images acquired from diverse viewpoints \citep{torii_are_2021, humenberger_investigating_2022}, the selection of reference candidates has a decisive influence on pose accuracy, as Section~\ref{exp_reference_candidates} shows.

\subsubsection{Feature extraction and matching}\label{m_image_matching}
We use sparse local features, which can be robustly re-identified under challenging conditions \citep{detone_superpoint_2018, tyszkiewicz_disk_2020, zhao_aliked_2023, potje_xfeat_2024, chen_rdd_2025}. Dense matchers outperform them under low texture, large viewpoint changes and difficult lighting \citep{jiang_imloc_2026, bonilla_mismatched_2025, panek_guide_2026}, but impose substantially higher computational and storage costs \citep{edstedt_roma_2024}, which becomes critical in large-scale, high-resolution settings.

Current sparse local features are typically trained on images of approximately 1 MP or less, e.g., ALIKED \citep{zhao_aliked_2023} on 800×800 and DISK \citep{tyszkiewicz_disk_2020} on 1024×1024 pixels. To apply these methods to high‐resolution images, we adopt a tiled feature extraction strategy, following \citet{maiwald_solving_2023} and \citet{morelli_deep-image-matching_2024}. Because features close to tile borders are often discarded for lack of context, gaps can appear near tile borders and introduce systematic effects. To mitigate this effect, tiles are arranged to overlap by a predefined number of pixels, as in \citet{morelli_deep-image-matching_2024}. After feature extraction, we reassemble the tiled outputs into a single feature set and apply non-maximum suppression (NMS) with a radius of 1 pixel to remove duplicate detections in the overlapping image regions. Finally, the query and reference images are matched exhaustively to obtain 2D–2D correspondences. 

\subsubsection{Local SfM-model reconstruction and pose estimation}\label{m_sfm_pose_estimation}
To estimate the pose of the query image, we first reconstruct a locally consistent SfM model by triangulating 3D points from geometrically verified 2D–2D correspondences between the reference images. In this step, both the camera intrinsics and the poses of the reference images are kept fixed. The 2D–2D correspondences between the query image and the reference images are subsequently examined to determine whether they are associated with reconstructed 3D points, thereby yielding a set of 2D–3D correspondences. Following the findings of \citet{vultaggio_perspective-n-point_2025}, we estimate the query image's pose using a PnP solver within a RANSAC framework, followed by non-linear pose refinement. In this work, only undistorted images are used (Section~\ref{m_dataset_query_sequences}). Therefore, the intrinsics are kept fixed during pose estimation.

\section{Experiments}\label{experiments} 
We perform four experiments, each addressing one of the sub-questions (i)--(iv) stated in Section~\ref{introduction}. Table~\ref{tbl_exp_config} summarizes the configuration of each experiment.

\begin{table*}[ht]
\centering
\caption{Configurations of the experiments: query and reference imagery, image retrieval (IR) method, position and heading pre-filtering, local features, and image resolution levels.}\label{tbl_exp_config}
\setlength{\tabcolsep}{4pt}
\resizebox{\textwidth}{!}{
\begin{tabular}{lllllll}
\toprule
Experiment & Reference imagery & Query imagery & IR method & Pre-filtering & Local features & Resolution \\ \midrule
(i) Sec.~\ref{exp_reference_candidates} & 2023, all cameras & 2025 panorama, front head & six methods & five strategies & SuperPoint & 2048 \\
(ii) Sec.~\ref{exp_high_res_imagery} & 2023 stereo & 2025 mono front & EigenPlaces (512) & $\pm$15\,m / $\pm$90$^\circ$ & six features & 1024--full \\
(iii) Sec.~\ref{exp_consumer_query} & 2023 / 2025 / both & consumer cameras & MegaLoc & $\pm$100\,m$^{a}$ / -- & five features & 2048 \\
(iv) Sec.~\ref{exp_local_vs_global} & 2025 & consumer cameras & MegaLoc & $\pm$100\,m$^{a}$ / -- & ALIKED & 1024 \\
\bottomrule
\end{tabular}}
\smallskip
{\footnotesize $^{a}$ Centered on one manually selected map coordinate per scene, identical for all query images of that scene.\par}
\end{table*}

\subsection{Evaluation protocol and metrics}\label{exp_eval_protocol_metrics}
For all experiments, we evaluate the performance of visual localization by the pose errors. Following common practice, we report the percentage of query images that are localized within specific error thresholds \citep{sattler_benchmarking_2018}. We compute the translation $\epsilon_{t}$ and rotation error $\epsilon_{R}$ according to Equations \eqref{formula_e_t} and \eqref{formula_e_r} and apply seven error thresholds: [0.01, 0.025, 0.05, 0.1, 0.2, 0.5, 1] meters for translation and [0.1, 0.25, 0.5, 1, 2, 5, 10] degrees for rotation. In addition, we use three error categories: accurate (0.05 m, 0.5$^\circ$), moderate (0.1 m, 1$^\circ$), and low (0.2 m, 2$^\circ$) for concise reporting. Following \citet{jin_image_2021}, we set $\epsilon_{t}$ and $\epsilon_{R}$ to infinity for images for which localization failed. Where appropriate, we also report the median translation and rotation errors.
\begin{equation}\label{formula_e_t}
\epsilon_{t} = \|t_{est}-t_{gt}\|_{2}
\end{equation}
\begin{equation}\label{formula_e_r}
\epsilon_{R} = \arccos\left(\frac{trace(R_{gt}^{-1}R_{est}) - 1}{2}\right)
\end{equation}

All pose errors reported in this work are computed with respect to the ground-truth poses and therefore describe the accuracy relative to the reference imagery. The absolute accuracy additionally depends on the georeferencing of the reference imagery (Section~\ref{dis_accuracy_potential}) and has to be considered if absolute pose errors are of interest. The strictest translation threshold of 0.01\,m is of the same order as the uncertainty of the ground truth: the two mobile mapping campaigns, whose images serve as queries in Sections~\ref{exp_reference_candidates} and~\ref{exp_high_res_imagery}, are co-registered with a 3D RMSE of 5 mm at independent tie points (Section~\ref{m_dataset_reference_imgs}), and the positions of the consumer queries have an uncertainty of 6–9 mm (Section~\ref{m_dataset_query_sensitivity}). At this threshold, part of the measured error stems from the ground truth itself, so the reported percentages are a conservative estimate. The rotation thresholds are not affected, as the orientation uncertainty of the ground truth is well below 0.1$^\circ$. 

Images acquired in a sequence are not independent samples, as they share the static scene content and differ mainly in dynamic elements such as moving vehicles. The five scenes were moreover selected deliberately rather than drawn at random. We therefore report the results per scene and camera and base our conclusions on differences that appear consistently across the conditions varied in each experiment, such as scenes, cameras, image resolutions or retrieval methods. For each difference, we report in how many of these combinations it is observed.

\subsection{Implementation details}\label{exp_implementation_details}
The implementation of our visual localization pipeline adopts a modular design, composed of many different software modules. Unless otherwise specified, the default values for each method are used.

For reference candidate selection, we employ six IR methods: NetVLAD \citep{arandjelovic_netvlad_2016}, CosPlace \citep{berton_rethinking_2022}, EigenPlaces \citep{berton_eigenplaces_2023}, AnyLoc \citep{keetha_anyloc_2024}, DINOv2 SALAD \citep{izquierdo_optimal_2024}, and MegaLoc \citep{berton_megaloc_2025}, using descriptor dimensions of 512 and 2048 for CosPlace and EigenPlaces. For AnyLoc, we use VLAD aggregation of DINOv2 ViT-G/14 features with 32 clusters and the precomputed urban vocabulary of \citet{keetha_anyloc_2024}. Following \citet{berton_megaloc_2025}, images are resized to 322$\times$322 pixels for the vision transformer-based methods (AnyLoc, DINOv2 SALAD, MegaLoc), whereas for NetVLAD, CosPlace, and EigenPlaces the longer image side is scaled to 322 pixels, preserving the aspect ratio. For the $k$-nearest-neighbor search, we set $k = 10$, which offers a good trade-off between efficiency and accuracy \citep{humenberger_investigating_2022}.

We use SuperPoint \citep{detone_superpoint_2018}, DISK \citep{tyszkiewicz_disk_2020}, ALIKED \citep{zhao_aliked_2023}, XFeat \citep{potje_xfeat_2024} and RDD \citep{chen_rdd_2025} as local feature representations and SIFT \citep{lowe_distinctive_2004}, implemented in OpenCV \citep{opencv_library}, as a baseline. According to their original publications, SuperPoint, DISK and XFeat detect keypoints with pixel-level localization accuracy, while SIFT, ALIKED and RDD provide sub-pixel-accurate keypoint locations.

Within our tiled feature extraction module, we use tiles of 1024$\times$1024 pixels, overlapping by 50 pixels, for all feature representations except SuperPoint, for which we use 1600$\times$1600 pixels following the image resolution used by \citet{sarlin_superglue_2020}. We define four image resolution levels, 1024, 2048, 4096 and full resolution, by scaling the longer image side to the chosen level while preserving the aspect ratio. The number of keypoints per image is limited to 4096 and 6144 at the resolution levels 1024 and 2048, respectively, and to 1024 keypoints per megapixel at 4096 pixels and full resolution. Apart from the number of keypoints, the default settings of each feature extraction method are used.

2D--2D image correspondences are obtained with LightGlue \citep{lindenberger_lightglue_2023}, a widely used state-of-the-art baseline \citep{image-matching-challenge-2025}, except for XFeat, matched with LighterGlue, and SIFT, matched by mutual nearest neighbor (MNN). To achieve the highest accuracy and comparability across different feature representations, we disable depth confidence and point pruning in LightGlue and LighterGlue.

For 3D point triangulation and absolute pose estimation, we use COLMAP \citep{schonberger_structure--motion_2016} with its default settings, with the following exceptions. The image poses of the reference images are kept fixed during triangulation and the camera intrinsics are not refined, as all images are undistorted (Section~\ref{m_dataset_query_sequences}). We also enable two-view tracks to retain points observed in only two of the retrieved images, since each local model is built from few images. The query pose is estimated with COLMAP's LO-RANSAC-based estimator \citep{chum_locally_2003}, which uses a minimal P3P solver and EPnP \citep{lepetit_epnp_2009} for local optimization, with a reprojection-error threshold of 2 pixels, followed by non-linear pose refinement.  Only for the comparison with HLoc in Section~\ref{exp_local_vs_global}, we use HLoc's and COLMAP's default threshold of 12 pixels for both pipelines.

\subsection{Investigation of reference candidate selection}\label{exp_reference_candidates}

\begin{table*}[ht]
\centering
\caption{Effect of reference candidate selection: percentage of poses within the accurate (acc: 0.05\,m, 0.5$^\circ$), moderate (mod: 0.1\,m, 1$^\circ$) and low (0.2\,m, 2$^\circ$) thresholds per pre-filtering strategy ($\pm$position / $\pm$heading) and IR method. Queries: 2025 panorama front head; reference: 2023 campaign; SuperPoint at 2048 pixels. Best, second- and third-best results: \textbf{bold}, \uline{underlined}, \uwave{wavy underlined}.}\label{table_eval_global_desc}
\resizebox{\textwidth}{!}{
\begin{tabular}{lrccccc}
\toprule
 \multirow{2}{*}{IR method} &\multirow{2}{*}{\shortstack{Desc\\dim}} & -- / -- & $\pm$100 m / -- & $\pm$100 m / $\pm$90$^\circ$ & $\pm$15 m / -- & $\pm$15 m / $\pm$90$^\circ$ \\ 
 && acc / mod / low & acc / mod / low & acc / mod / low & acc / mod / low & acc / mod / low\\ \midrule
NetVLAD & 4096&52.1 / 66.5 / 70.2&60.8 / 77.1 / 81.7&64.3 / 82.8 / 88.0&71.2 / 88.7 / 92.3&77.2 / 94.0 / 97.0 \\
\multirow{2}{*}{CosPlace} &512 & 79.5 / 87.4 / 89.8&82.8 / 91.2 / 94.0&84.1 / 92.4 / 95.6&89.3 / 95.1 / 97.3&89.6 / 95.9 / \uwave{98.1} \\
&2048&81.0 / 87.9 / 90.4&84.9 / 91.8 / 94.1&86.4 / 92.8 / 95.3&89.2 / 95.3 / 97.2&90.4 / 96.1 / 97.9 \\
\multirow{2}{*}{EigenPlaces} &512 &84.0 / 91.7 / 94.0&86.1 / 93.9 / 96.1&86.5 / 94.3 / 96.7&90.2 / 96.2 / \uwave{97.8}&90.3 / 96.2 / 98.0 \\
&2048&\uwave{84.7} / \uwave{92.2} / \uwave{94.8}&\uwave{86.3} / \uwave{94.1} / \uwave{96.8}&\uwave{87.2} / \uwave{94.8} / \uwave{97.2}&\uwave{90.3} / \uwave{96.3} / \uline{97.9}&90.4 / \uwave{96.4} / \uwave{98.1} \\
AnyLoc &49152&78.5 / 86.2 / 88.3&79.6 / 87.3 / 89.7&84.3 / 92.5 / 95.1&87.3 / 93.3 / 95.6&\uwave{90.7} / \uwave{96.4} / 97.9 \\
DINOv2 SALAD &8448 &\uline{85.9} / \uline{93.3} / \uline{95.5}&\uline{87.6} / \uline{94.8} / \uline{97.0}&\uline{87.8} / \uline{95.1} / \uline{97.3}&\uline{91.2} / \textbf{96.6} / \textbf{98.3}&\uline{91.1} / \uline{96.5} / \uline{98.4} \\
MegaLoc&8448 &\textbf{90.3} / \textbf{95.9} / \textbf{97.6}&\textbf{90.3} / \textbf{96.3} / \textbf{98.3}&\textbf{90.4} / \textbf{96.0} / \textbf{98.2}&\textbf{91.5} / \uline{96.4} / \textbf{98.3}&\textbf{91.7} / \textbf{96.7} / \textbf{98.6} \\ \bottomrule
\end{tabular}}
\end{table*}

This experiment evaluated the reference candidate selection strategy described in Section~\ref{m_ref_cand_selection} by combining five pre-filtering strategies with the selected IR methods (Table~\ref{tbl_exp_config}).

As a baseline, we omitted the pre-filtering step and retrieved the 10 most similar images from the entire database. For position-based filtering, we used search regions of $\pm$100\,m and $\pm$15\,m. Each of these thresholds was combined with either no heading constraint or a heading threshold of $\pm$90$^\circ$ (i.e., a 180$^\circ$ acceptance window). Since smartphones typically achieve GNSS accuracies of several meters, $\pm$100\,m represents an upper bound that should be met under almost all practical conditions, whereas $\pm$15\,m is a lower bound defined by the spatial density of our reference data and the aim of reconstructing a local SfM model.

The filters were applied to the poses from integrated georeferencing, i.e., to nearly exact priors. Imperfect priors are nevertheless covered, as the configuration without pre-filtering corresponds to a missing position prior and the configurations without heading constraint correspond to a missing heading. A position error within the search region is expected to yield results between those of the unfiltered and the filtered configurations, whereas a larger error may exclude the correct reference images and cause the localization to fail. The search region should therefore be chosen conservatively with respect to the expected accuracy of the prior.

Pre-filtering improves pose accuracy in the accurate category in 31 of the 32 combinations of IR method and pre-filtering strategy, and leaves it unchanged in the remaining one (Table~\ref{table_eval_global_desc}), but the gain depends heavily on the unfiltered baseline. In the accurate category, the weak NetVLAD gains 25.1 percentage points (pp), the already strong MegaLoc only 1.4\,pp. Position-based filtering yields large and consistent improvements. Heading-based filtering adds only a marginal benefit for the stronger descriptors, but up to 6.0\,pp for NetVLAD and 4.7\,pp for AnyLoc. MegaLoc achieves the highest percentage in the accurate category for all pre-filtering strategies, and its lead is especially pronounced without pre-filtering. With the strictest pre-filtering ($\pm$15\,m / $\pm$90$^\circ$), the IR methods differ by only about 2\,pp (NetVLAD excepted), which shows that pre-filtering can effectively compensate for weaker descriptor expressiveness. The vision transformer-based methods MegaLoc and DINOv2 SALAD produce longer but more expressive descriptors than the convolutional neural network (CNN)-based methods NetVLAD, CosPlace, and EigenPlaces. The 512-dimensional descriptors of CosPlace and EigenPlaces perform only slightly worse than their 2048-dimensional counterparts, and for EigenPlaces almost equivalently. The clear exception is AnyLoc, which despite its transformer backbone and very long descriptor performs worse than DINOv2 SALAD and MegaLoc throughout and, without heading filter, also worse than CosPlace and EigenPlaces. This agrees with other evaluations on street-level data \citep{berton_megaloc_2025, izquierdo_optimal_2024, haitz_evaluation_2026}. AnyLoc aggregates features of a frozen DINOv2 backbone with an unsupervised vocabulary and no task-specific training \citep{keetha_anyloc_2024}, whereas the other methods are trained on datasets that include street-level imagery similar to ours.

\subsection{Effects of high-resolution images on visual localization}\label{exp_high_res_imagery}

\begin{table*}[ht]
\centering
\caption{Effect of image resolution and local features: percentage of poses within seven error thresholds and median pose errors. Queries: 2025 mono front camera; reference: 2023 stereo cameras. \uuline{Double underline}: best overall per threshold; \uwave{wavy underline}: best resolution per feature; colored entries: best feature per resolution (\tcb{1024}{1024}, \tcb{2048}{2048}, \tcb{4096}{4096}, and \tcb{full}{full}).}\label{tbl_exp_image_res_feats}
\begin{tabular}{lc*{2}{S[table-format=1.3]}*{7}{S[table-format=2.1]}}
\toprule
\multirow{2}{*}{Method} & {Image} & \multicolumn{2}{c}{Med. pose error} &{0.01m}& {0.025m} &{0.05m} &{0.1m}&{0.2m}&{0.5m} &{1m} \\
&{res} & {$\epsilon_{t}$ [m]} &{$\epsilon_{R}$ [$^\circ$]} & {0.1$^\circ$} & {0.25$^\circ$} & {0.5$^\circ$} & {1$^\circ$} & {2$^\circ$} & {5$^\circ$} & {10$^\circ$} \\ \midrule
\multirow{4}{*}{SIFT}
& 1024&\uwave{0.114}&\uwave{0.213}&7.3&20.3&33.6&\uwave{47.4}&\uwave{58.1}&\uwave{64.8}&\uwave{66.7} \\
& 2048&0.124&0.215&13.0&\uwave{26.9}&\uwave{37.4}&46.8&55.0&60.1&61.6 \\
& 4096&0.873&1.849&\uwave{14.1}&24.7&32.2&38.6&44.1&48.7&50.3 \\
& full&{-}$^{a}$&{-}$^{a}$&13.6&22.5&28.8&34.8&40.4&45.2&46.8 \\
\midrule
\multirow{4}{*}{SuperPoint}
& 1024&0.022&0.051&17.4&56.4&79.4&91.5&96.5&98.4&98.7 \\
& 2048&0.013&\tcb{2048}{0.031}&35.9&78.1&92.2&97.2&\uwave{98.5}&\uwave{98.9}&\uwave{99.0} \\
& 4096&\uwave{0.011}&\tcb{4096}{0.027}&45.9&82.2&93.7&\uwave{97.5}&98.2&98.6&98.7 \\
& full&\uwave{0.011}&\tcb{full}{\uuline{0.026}}&\uwave{47.1}&\uwave{82.8}&\uwave{94.0}&\uwave{97.5}&98.3&98.6&98.7 \\
\midrule
\multirow{4}{*}{DISK}
& 1024&0.021&\tcb{1024}{0.047}&24.4&55.5&76.0&89.0&94.3&\uwave{97.0}&\uwave{97.6} \\
& 2048&\uwave{0.012}&\uwave{0.033}&\uwave{41.1}&\uwave{72.3}&\uwave{85.6}&\uwave{92.2}&\uwave{94.8}&96.1&96.6 \\
& 4096&0.015&0.034&38.3&62.9&75.5&81.9&84.8&86.5&86.9 \\
& full&0.018&0.035&32.9&58.5&71.0&77.4&79.4&80.6&81.0 \\
\midrule
\multirow{4}{*}{ALIKED}
& 1024&0.019&0.049&21.1&61.8&81.4&92.5&96.7&98.3&98.7 \\
& 2048&0.012&0.034&41.4&80.0&92.8&\uwave{97.0}&\uwave{98.3}&\uwave{98.9}&\uwave{99.0} \\
& 4096&\uwave{0.011}&0.030&\uwave{46.3}&\uwave{81.6}&\uwave{93.2}&96.8&97.8&98.2&98.3 \\
& full&0.012&\uwave{0.029}&41.7&78.2&91.7&95.7&97.2&97.7&97.7 \\
\midrule
\multirow{4}{*}{XFeat}
& 1024&0.045&0.080&5.4&26.8&53.7&76.4&89.2&95.8&97.4 \\
& 2048&0.026&0.046&14.1&48.7&74.3&88.6&\uwave{94.2}&\uwave{96.9}&\uwave{97.5} \\
& 4096&0.019&0.036&23.1&61.7&\uwave{82.6}&\uwave{90.1}&93.2&94.6&95.1 \\
& full&\uwave{0.018}&\uwave{0.035}&\uwave{28.9}&\uwave{61.9}&79.0&86.6&89.9&91.2&91.6 \\
\midrule
\multirow{4}{*}{RDD}
& 1024&\tcb{1024}{0.012}&\tcb{1024}{0.047}&\tcb{1024}{39.0}&\tcb{1024}{76.1}&\tcb{1024}{89.2}&\tcb{1024}{95.0}&\tcb{1024}{97.5}&\tcb{1024}{98.8}&\tcb{1024}{98.9} \\
& 2048&\tcb{2048}{\uuline{0.009}}&0.032&\tcb{2048}{\uuline{53.4}}&\tcb{2048}{\uuline{85.9}}&\tcb{2048}{94.8}&\tcb{2048}{97.5}&\tcb{2048}{98.7}&\tcb{2048}{99.0}&\tcb{2048}{99.1} \\
& 4096&\tcb{4096}{0.010}&0.030&\tcb{4096}{48.5}&\tcb{4096}{83.3}&\tcb{4096}{94.2}&\tcb{4096}{97.6}&\tcb{4096}{98.5}&\tcb{4096}{99.0}&\tcb{4096}{99.0} \\
& full&\tcb{full}{0.010}&\uwave{0.029}&\tcb{full}{48.4}&\tcb{full}{84.7}&\tcb{full}{\uuline{95.2}}&\tcb{full}{\uuline{98.2}}&\tcb{full}{\uuline{98.9}}&\tcb{full}{\uuline{99.1}}&\tcb{full}{\uuline{99.2}} \\

\bottomrule
\end{tabular}

\smallskip
{\footnotesize $^{a}$ Undefined, as fewer than 50\,\% of the images were successfully localized.\par}
\end{table*}

In general, higher image resolutions yield more accurate image observations. In our visual localization pipeline, the image observations affect the accuracy of the estimated pose of the query image in two ways, directly through the keypoints extracted from the query image and indirectly through the accuracy of the reconstructed 3D point coordinates.

To evaluate the impact of image resolution on the final image poses, we tested the six local feature representations at the four image resolution levels, using the high-resolution images of the 2025 monocular front camera as queries and the images of the three 2023 stereo systems as reference (Table~\ref{tbl_exp_config}). We excluded the panorama camera, as its substantially lower spatial resolution (Table~\ref{mms_sensor_setup}) would introduce heterogeneous keypoint localization accuracies within an otherwise uniformly high-resolution setting. As the strict $\pm$15 m / $\pm90^\circ$ pre-filtering makes the choice of IR method nearly irrelevant (Section~\ref{exp_reference_candidates}), we used the compact 512-dimensional EigenPlaces descriptor.

\begin{figure}[ht]
\includegraphics[width=\linewidth,keepaspectratio]{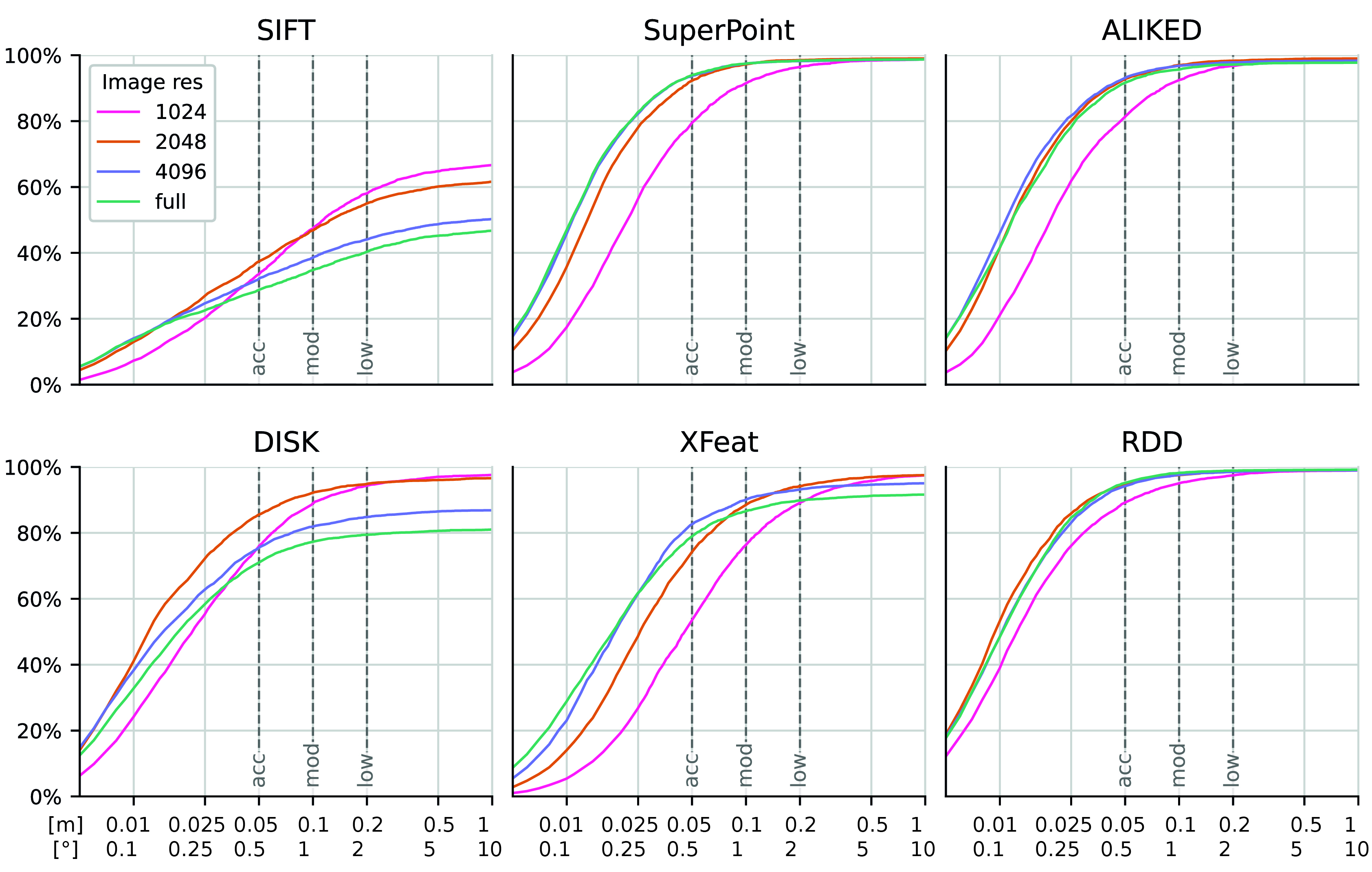}
\caption{Percentage of poses with translation error $\epsilon_t$ and rotation error $\epsilon_R$ below the value on the logarithmic x-axis (0.0005--1\,m and 0.005--10$^\circ$), for six local features (SIFT, SuperPoint, DISK, ALIKED, XFeat and RDD) at four image resolution levels (\textcolor{1024}{1024}, \textcolor{2048}{2048}, \textcolor{4096}{4096} and \textcolor{full}{full}).}\label{fig_exp_img_res_feat}
\end{figure}

Since both query and reference images were recorded by MMS whose trajectories are constrained by lanes, strong viewpoint changes were largely avoided. This matters because many local feature representations are neither scale- nor rotation-invariant, with the lack of rotation invariance having the considerably stronger adverse effect \citep{tyszkiewicz_disk_2020, zhao_aliked_2023}. For the same reason, we applied the resolution level to both query and reference imagery, which avoids a scale difference between them. The reported results therefore reflect the resolution of the scene representation together with that of the query imagery. The setup thus isolates the effect of image resolution, while changes in vegetation and weather, ongoing construction work and differing camera intrinsics still reflect realistic conditions.

The effect of image resolution is most pronounced at the two strictest thresholds, 0.01\,m / 0.1$^\circ$ and 0.025\,m / 0.25$^\circ$ (Table~\ref{tbl_exp_image_res_feats}, Figure~\ref{fig_exp_img_res_feat}). Increasing the resolution from 1024 to 2048 pixels improves all learned local features, reducing their median translation errors by 25--43\,\% and their median rotation errors by 30--43\,\%. Beyond 2048 pixels, only SuperPoint benefits from higher resolution throughout, though with diminishing returns, whereas SIFT shows no consistent optimum. DISK exhibits a resolution-dependent optimum at 2048 pixels, whereas for ALIKED, XFeat and RDD the optimum additionally depends on the error threshold, with ALIKED peaking at 4096 pixels in the high-accuracy regime, i.e., at the thresholds of 0.05\,m / 0.5$^\circ$ and below, and at 2048 pixels at the coarser thresholds. RDD attains the highest percentage of all configurations at all seven error thresholds, by 6.3 and 3.1 pp at the two strictest thresholds and by less than 1 pp from 0.1\,m / 1$^\circ$ onward. Note that the number of extracted keypoints grows with the resolution (Section~\ref{exp_implementation_details}) and that the RANSAC threshold is fixed in pixels, so the results reflect these combined settings rather than the resolution alone.

\subsection{Visual localization performance with consumer-grade query images}\label{exp_consumer_query}

We then assessed the accuracy potential using the query images acquired with consumer devices. Unlike the previous experiment, these query images include rotated and scaled image content due to strong viewpoint changes, as well as close-up images with very little visual context. We used three distinct reference configurations: the 2023 campaign, the 2025 campaign and both campaigns combined. Combining both campaigns simulates a reference map built from multiple epochs, as occurs when reference data is added rather than replaced during updates.

We omitted SIFT because of its inferior performance in Section~\ref{exp_high_res_imagery} and used a resolution of 2048 pixels. For reference candidate selection, we used MegaLoc, the strongest IR method, with a position-based pre-filter of $\pm$100\,m. For each scene, the search region was centered on a single coordinate selected manually on a map, so that the prior was independent of the ground truth and identical for all query images of a scene.

\begin{figure*}[ht]
\centering
\includegraphics[width=0.85\linewidth,height=0.95\textheight,keepaspectratio]{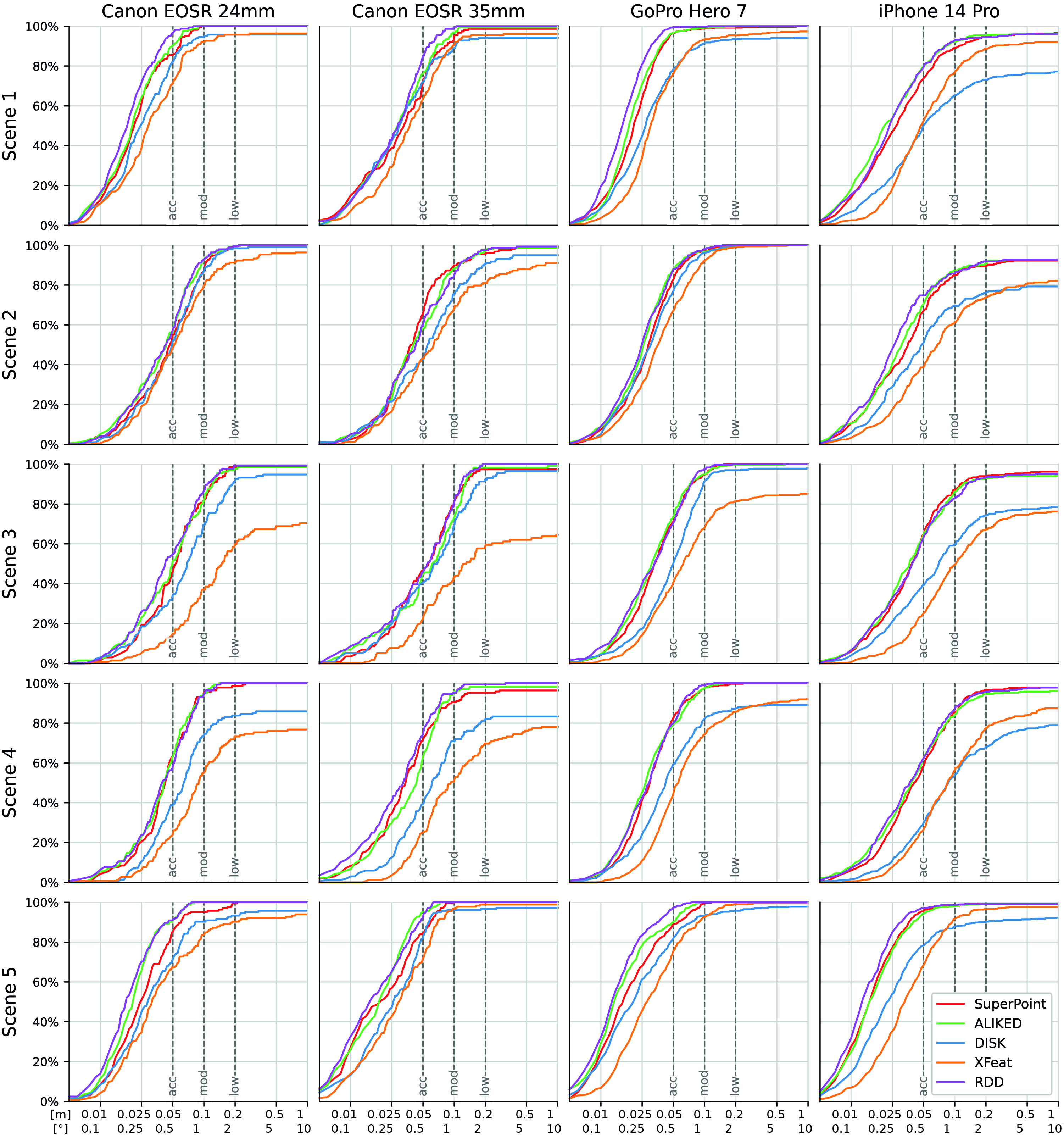} 
\caption{Localization performance per scene and camera against the 2025 reference imagery: percentage of poses with $\epsilon_t$ and $\epsilon_R$ below the value on the logarithmic x-axis (0.0005--1\,m and 0.005--10$^\circ$). MegaLoc with $\pm$100\,m position pre-filtering; image resolution 2048 pixels.}\label{fig_exp_local_scenes_feats_scene_cam}
\end{figure*}

Tables~\ref{tbl_exp_local_scenes_agg_cat_2023}, \ref{tbl_exp_local_scenes_agg_cat_2025} and~\ref{tbl_exp_local_scenes_agg_cat_2023_2025} report the results aggregated over the five scenes for the 2023, 2025 and combined reference data, and Table~\ref{tbl_exp_local_scenes_agg_median_pose_error} compares the median pose errors. Since current local feature representations exhibit environment-dependent performance due to limited training data \citep{bonilla_mismatched_2025}, Figure~\ref{fig_exp_local_scenes_feats_scene_cam} additionally shows the results per scene and camera for the 2025 reference data, which is subject to the smallest long-term changes and therefore allows the clearest comparison of the local feature representations.

\begin{table}[ht]
\centering
\caption{Localizing consumer cameras against the 2023 reference campaign: percentage of poses within seven error thresholds per camera and local feature (S: SuperPoint, D: DISK, A: ALIKED, X: XFeat, R: RDD). MegaLoc with $\pm$100\,m position pre-filtering; 2048 pixels. Best per camera in bold.}\label{tbl_exp_local_scenes_agg_cat_2023}
\setlength{\tabcolsep}{2.5pt}
\begin{tabular}{l l*{7}{S[table-format=2.1]}} 
\toprule
 \multirow[c]{2}{*}{\rotatebox[origin=c]{90}{Cam}} & \multirow[c]{2}{*}{\rotatebox[origin=c]{90}{Feat}} & {0.01m}& {0.025m} &{0.05m} &{0.1m}&{0.2m}&{0.5m} &{1m } \\
 && {0.1$^\circ$} & {0.25$^\circ$} & {0.5$^\circ$} & {1$^\circ$} & {2$^\circ$} & {5$^\circ$} & {10$^\circ$}\\ 
 \midrule
 \multirow[c]{5}{*}{\rotatebox[origin=c]{90}{Canon 24mm}}
&S& \bfseries 4.7&\bfseries 28.1&58.8&85.8&95.3&97.5&97.8 \\
&D& 2.0&12.6&33.0&55.6&69.0&74.4&75.7 \\
&A& 2.1&21.8&53.4&80.8&91.8&94.3&94.5 \\
&X& 1.2&7.1&22.2&43.4&54.8&61.7&64.7 \\
&R& 3.1&27.0&\bfseries 59.3&\bfseries 86.0&\bfseries 95.9&\bfseries 98.0&\bfseries 98.3 \\ \midrule
 \multirow[c]{5}{*}{\rotatebox[origin=c]{90}{Canon 35mm}}
&S& 10.4&32.4&59.5&\bfseries 80.6&90.1&94.0&95.1 \\
&D& 3.6&14.5&27.0&44.9&51.0&56.8&58.0 \\
&A& 8.8&28.6&55.1&77.2&89.1&94.1&95.2 \\
&X& 1.9&9.5&22.4&35.6&46.2&52.3&55.5 \\
&R& \bfseries 10.9&\bfseries 37.5&\bfseries 60.7&79.9&\bfseries 92.8&\bfseries 97.0&\bfseries 98.3 \\ \midrule
 \multirow[c]{5}{*}{\rotatebox[origin=c]{90}{GoPro}}
&S& 6.2&37.3&71.8&91.2&96.9&98.5&99.2 \\
&D& 3.5&20.2&47.5&69.7&79.4&85.5&87.1 \\
&A& 3.9&30.0&66.7&88.6&95.8&97.3&98.1 \\
&X& 1.8&13.7&35.6&62.1&75.3&83.1&85.5 \\
&R& \bfseries 6.4&\bfseries 40.2&\bfseries 72.3&\bfseries 92.9&\bfseries 98.4&\bfseries 99.5&\bfseries 99.6 \\ \midrule
 \multirow[c]{5}{*}{\rotatebox[origin=c]{90}{iPhone}}
&S& 5.0&\bfseries 26.4&\bfseries 50.1&\bfseries 72.4&\bfseries 85.4&\bfseries 90.0&\bfseries 92.1 \\
&D& 1.8&9.3&19.3&33.2&44.4&50.2&52.9 \\
&A& \bfseries 5.1&24.5&48.1&69.4&81.2&86.3&87.9 \\
&X& 0.8&6.0&18.1&36.4&54.9&68.0&72.5 \\
&R& 4.9&24.7&45.0&67.3&81.1&87.2&90.2 \\
\bottomrule
\end{tabular}
\end{table}

\begin{table}[ht]
\centering
\caption{Localizing consumer cameras against the 2025 reference campaign: percentage of poses within seven error thresholds per camera and local feature (S: SuperPoint, D: DISK, A: ALIKED, X: XFeat, R: RDD). MegaLoc with $\pm$100\,m position pre-filtering; 2048 pixels. Best per camera in bold.}\label{tbl_exp_local_scenes_agg_cat_2025}
\setlength{\tabcolsep}{2.5pt}
\begin{tabular}{l l *{4}{S[table-format=2.1]}*{3}{S[table-format=3.1]}} 
\toprule
 \multirow[c]{2}{*}{\rotatebox[origin=c]{90}{Cam}} & \multirow[c]{2}{*}{\rotatebox[origin=c]{90}{Feat}} & {0.01m}& {0.025m} &{0.05m} &{0.1m}&{0.2m}&{0.5m} &{1m } \\
 && {0.1$^\circ$} & {0.25$^\circ$} & {0.5$^\circ$} & {1$^\circ$} & {2$^\circ$} & {5$^\circ$} & {10$^\circ$}\\ 
\midrule
 \multirow[c]{5}{*}{\rotatebox[origin=c]{90}{Canon 24mm}}
&S& 6.2&35.6&67.5&92.0&99.1&99.8&99.8 \\
&D& 5.1&29.7&56.8&83.9&93.1&94.6&94.6 \\
&A& 8.1&42.1&71.0&94.4&99.4&99.8&99.8 \\
&X& 3.5&21.7&46.9&71.5&83.5&87.1&88.0 \\
&R& \bfseries 8.7&\bfseries 44.6&\bfseries 72.2&\bfseries 95.0&\bfseries 99.8&\bfseries 99.9&\bfseries 99.9 \\ \midrule
\multirow[c]{5}{*}{\rotatebox[origin=c]{90}{Canon 35mm}}
&S& 12.1&36.7&69.9&92.1&97.4&98.3&98.3 \\
&D& 7.6&29.7&56.5&81.1&91.1&93.1&93.1 \\
&A& 12.0&38.1&69.5&92.0&98.6&99.0&99.1 \\
&X& 4.9&21.2&47.0&71.8&82.4&85.7&87.1 \\
&R& \bfseries 15.2&\bfseries 41.4&\bfseries 73.7&\bfseries 92.9&\bfseries 99.4&\bfseries 99.9&\bfseries 99.9 \\ \midrule
 \multirow[c]{5}{*}{\rotatebox[origin=c]{90}{GoPro}}
&S& 10.9&49.8&84.7&97.6&99.7&99.9&\bfseries 100.0 \\
&D& 8.7&38.3&69.7&90.9&94.6&95.6&95.8 \\
&A& 11.6&56.7&85.7&97.7&99.8&99.9&\bfseries 100.0 \\
&X& 3.2&27.5&61.4&84.3&92.2&94.3&95.0 \\
&R& \bfseries 15.3&\bfseries 60.1&\bfseries 87.4&\bfseries 98.9&\bfseries 100.0&\bfseries 100.0&\bfseries 100.0 \\ \midrule
 \multirow[c]{5}{*}{\rotatebox[origin=c]{90}{iPhone}}
&S& 13.8&46.4&73.6&89.6&\bfseries 95.4&\bfseries 96.7&\bfseries 96.9 \\
&D& 7.0&28.9&51.2&68.1&76.9&80.8&81.9 \\
&A& 14.6&49.5&76.0&90.2&95.1&96.0&96.2 \\
&X& 2.9&18.5&44.2&69.2&82.4&86.9&88.2 \\
&R& \bfseries 16.0&\bfseries 52.9&\bfseries 76.7&\bfseries 90.3&\bfseries 95.4&96.4&96.6 \\
\bottomrule

\end{tabular}
\end{table}

\begin{table}[ht]
\centering
\caption{Localizing consumer cameras against the combined 2023 and 2025 reference campaigns: percentage of poses within seven error thresholds per camera and local feature (S: SuperPoint, D: DISK, A: ALIKED, X: XFeat, R: RDD). MegaLoc with $\pm$100\,m position pre-filtering; image resolution 2048 pixels. Best per camera in bold.} \label{tbl_exp_local_scenes_agg_cat_2023_2025}
\setlength{\tabcolsep}{2.5pt}
\begin{tabular}{l l *{7}{S[table-format=2.1]}}
\toprule
 \multirow[c]{2}{*}{\rotatebox[origin=c]{90}{Cam}} & \multirow[c]{2}{*}{\rotatebox[origin=c]{90}{Feat}} & {0.01m}& {0.025m} &{0.05m} &{0.1m}&{0.2m}&{0.5m} &{1m } \\
 && {0.1$^\circ$} & {0.25$^\circ$} & {0.5$^\circ$} & {1$^\circ$} & {2$^\circ$} & {5$^\circ$} & {10$^\circ$}\\ \midrule
 \multirow[c]{5}{*}{\rotatebox[origin=c]{90}{Canon 24mm}}
&S& 5.0&29.1&62.8&\bfseries 88.0&\bfseries 94.3&\bfseries 96.0&\bfseries 97.3 \\
&D& 4.9&22.7&48.7&74.7&83.5&85.5&87.8 \\
&A& 5.9&33.5&61.8&85.6&91.3&92.8&94.4 \\
&X& 2.2&16.5&39.8&60.8&72.3&77.3&80.8 \\
&R& \bfseries 6.9&\bfseries 38.0&\bfseries 64.9&87.3&92.4&93.5&94.8 \\ \midrule
 \multirow[c]{5}{*}{\rotatebox[origin=c]{90}{Canon 35mm}}
&S& 10.4&29.9&63.3&88.2&93.7&94.5&96.1 \\
&D& 5.3&24.5&51.2&73.6&84.3&86.0&87.1 \\
&A& 9.8&31.7&64.2&88.2&92.3&93.3&94.5 \\
&X& 4.8&19.3&42.3&65.6&75.5&79.8&82.6 \\
&R& \bfseries 12.1&\bfseries 37.1&\bfseries 68.3&\bfseries 89.8&\bfseries 94.7&\bfseries 95.6&\bfseries 96.4 \\ \midrule
 \multirow[c]{5}{*}{\rotatebox[origin=c]{90}{GoPro}}
&S& 6.3&36.1&68.0&86.0&\bfseries 91.8&\bfseries 93.2&\bfseries 94.8 \\
&D& 4.8&26.2&51.4&68.8&76.7&81.2&83.9 \\
&A& 5.6&35.8&67.7&83.6&88.2&89.6&90.7 \\
&X& 2.1&16.4&39.7&63.6&75.6&80.8&83.3 \\
&R& \bfseries 8.2&\bfseries 42.9&\bfseries 71.6&\bfseries 86.5&90.1&91.4&92.5 \\ \midrule
 \multirow[c]{5}{*}{\rotatebox[origin=c]{90}{iPhone}}
&S& 8.9&32.7&55.3&\bfseries 71.5&\bfseries 81.1&\bfseries 86.1&\bfseries 88.8 \\
&D& 5.1&21.0&34.0&46.7&54.6&60.6&64.6 \\
&A& 11.7&37.1&\bfseries 57.8&71.2&78.5&83.2&85.7 \\
&X& 2.2&11.9&29.2&47.4&60.6&69.5&74.2 \\
&R& \bfseries 12.6&\bfseries 37.3&56.5&68.7&76.9&82.7&85.5 \\
\bottomrule

\end{tabular}
\end{table}

\begin{table}[ht]
\centering
\caption{Median translation ($\epsilon_t$) and rotation ($\epsilon_R$) errors per reference imagery, camera and local feature (S: SuperPoint, D: DISK, A: ALIKED, X: XFeat, R: RDD). MegaLoc with $\pm$100\,m position pre-filtering; image resolution 2048 pixels. Best entry per camera in bold.} \label{tbl_exp_local_scenes_agg_median_pose_error}
\setlength{\tabcolsep}{5pt}
\begin{tabular}{l l *{2}{S[table-format=1.3]} *{2}{S[table-format=1.3]}  *{2}{S[table-format=1.3]}}
\toprule
  \multirow[c]{2}{*}{\rotatebox[origin=c]{90}{Cam}} & \multirow[c]{2}{*}{\rotatebox[origin=c]{90}{Feat}} & \multicolumn{2}{c}{2023} &\multicolumn{2}{c}{2025}&\multicolumn{2}{c}{2023 \& 2025} \\
 && {$\epsilon_{t}$ [m]} &{$\epsilon_{R}$ [$^\circ$]} & {$\epsilon_{t}$ [m]} &{$\epsilon_{R}$ [$^\circ$]} & {$\epsilon_{t}$ [m]} &{$\epsilon_{R}$ [$^\circ$]}\\ \midrule 
 \multirow[c]{5}{*}{\rotatebox[origin=c]{90}{Canon 24mm}}
&S& 0.042&\bfseries 0.092&0.035&0.073&0.039&0.085 \\
&D& 0.081&0.143&0.043&0.078&0.052&0.097 \\
&A& 0.046&0.099&0.030&\bfseries 0.069&0.037&\bfseries 0.082 \\
&X& 0.147&0.249&0.054&0.104&0.071&0.124 \\
&R& \bfseries 0.041&0.096&\bfseries 0.029&0.070&\bfseries 0.034&\bfseries 0.082 \\ \midrule
 \multirow[c]{5}{*}{\rotatebox[origin=c]{90}{Canon 35mm}}
&S& 0.040&0.055&0.034&0.034&0.038&0.038 \\
&D& 0.179&0.142&0.042&0.040&0.049&0.045 \\
&A& 0.044&\bfseries 0.053&0.032&\bfseries 0.032&0.038&\bfseries 0.036 \\
&X& 0.280&0.362&0.054&0.058&0.061&0.061 \\
&R& \bfseries 0.037&0.054&\bfseries 0.031&\bfseries 0.032&\bfseries 0.035&0.037 \\ \midrule
 \multirow[c]{5}{*}{\rotatebox[origin=c]{90}{GoPro}}
&S& 0.032&0.086&0.025&0.063&0.033&0.087 \\
&D& 0.053&0.106&0.033&0.067&0.047&0.107 \\
&A& 0.036&0.087&0.022&\bfseries 0.060&0.033&\bfseries 0.082 \\
&X& 0.069&0.144&0.039&0.082&0.065&0.136 \\
&R& \bfseries 0.031&\bfseries 0.085&\bfseries 0.020&0.061&\bfseries 0.029&0.083 \\ \midrule
 \multirow[c]{5}{*}{\rotatebox[origin=c]{90}{iPhone}}
&S& \bfseries 0.050&\bfseries 0.094&0.028&0.056&0.042&0.084 \\
&D& 0.472&0.533&0.048&0.068&0.127&0.160 \\
&A& 0.052&0.098&0.025&\bfseries 0.053&\bfseries 0.037&\bfseries 0.078 \\
&X& 0.165&0.243&0.057&0.084&0.114&0.163 \\
&R& 0.058&0.104&\bfseries 0.023&\bfseries 0.053&0.038&0.083 \\
\bottomrule

\end{tabular}
\end{table}

Switching from the 2023 to the more recent 2025 reference data substantially improves localization accuracy. In the accurate category it yields a higher percentage for all 20 combinations of camera and feature representation (Tables~\ref{tbl_exp_local_scenes_agg_cat_2023} and~\ref{tbl_exp_local_scenes_agg_cat_2025}), and the median errors decrease accordingly (Table~\ref{tbl_exp_local_scenes_agg_median_pose_error}).
 
With both reference epochs, RDD, ALIKED and SuperPoint form a leading group. Against the 2025 reference data, RDD achieves the highest percentages at 0.025\,m / 0.25$^\circ$ and 0.05\,m / 0.5$^\circ$ and the lowest median translation errors for all four cameras, albeit by small margins of 2.5 to 3.4\,pp and 1 to 2\,mm (Tables~\ref{tbl_exp_local_scenes_agg_cat_2025} and~\ref{tbl_exp_local_scenes_agg_median_pose_error}). DISK and XFeat perform substantially worse with both epochs. For the 2025 reference data, they lag behind the best representation in the accurate category by 15 to 26\,pp and 25 to 33\,pp, respectively. At the coarsest thresholds the representations converge.
 
Per scene and camera (Figure~\ref{fig_exp_local_scenes_feats_scene_cam}), ALIKED, RDD and SuperPoint perform comparably and, in the accurate category, even the weakest of the three outperforms XFeat and matches or exceeds DISK in all 20 combinations of scene and camera. None of the three leads consistently, as RDD is best in 9 of the 20 combinations, ALIKED in 6 and SuperPoint in 5. The most pronounced drops for XFeat occur in scene~3, where moving or parked vehicles and reflective glass fa\c{c}ades produce features that cannot be matched reliably, and in scene~4, where bare branches yield fine-scale structures that are difficult to re-detect and repeated windows produce ambiguous correspondences, to which the comparatively compact descriptors of XFeat may be more susceptible.
 
Comparisons between the cameras must be interpreted with caution, as the sequences differ not only in the field of view, but also in trajectory, triggering, projection model and sensor resolution, and at 2048 pixels a wider field of view comes with a coarser angular resolution per pixel. With this caveat, the GoPro images (horizontal FoV of 122.6$^\circ$) yield the lowest median translation errors, whereas the Canon 35\,mm images (horizontal FoV: 53.3$^\circ$) achieve the lowest median rotation errors, about half of those of the Canon 24\,mm and the GoPro images (Table~\ref{tbl_exp_local_scenes_agg_median_pose_error}), consistent with their finer angular resolution.
 
With the combined reference data (Table~\ref{tbl_exp_local_scenes_agg_cat_2023_2025}), the median translation and rotation errors lie between those of the two single epochs for 19 and 18 of the 20 combinations, respectively (Table~\ref{tbl_exp_local_scenes_agg_median_pose_error}). At coarser thresholds, however, the combined data frequently perform worse than the 2023 data alone, for 13 of the 20 combinations at 1\,m / 10$^\circ$ (e.g., 94.8\,\% vs. 99.2\,\% for the GoPro with SuperPoint), indicating more localization failures. Possible reasons for this behavior and its implications for updating the reference data are discussed in Section~\ref{dis_reference_epoch}.

\subsection{Local vs. global SfM-based visual localization}\label{exp_local_vs_global}
According to \citet{panek_guide_2026}, global SfM-based visual localization methods are among the most accurate. To quantify the accuracy cost of building a local SfM model on the fly from georeferenced images, we compared our pipeline with HLoc \citep{sarlin_coarse_2019}, which we consider a representative method of this category.

We localized the query images of all four cameras against the 2025 reference data, using identical settings for both pipelines. These are: MegaLoc with $k = 10$, ALIKED features at 1024 pixels, and HLoc's default RANSAC threshold of 12 pixels instead of the 2 pixels used in the other experiments. For each scene, we selected the reference images within $\pm$100\,m of the same manually selected coordinate as in Section~\ref{exp_consumer_query} and triangulated a globally consistent SfM model from their exhaustive matches, as required by HLoc. For each query image, MegaLoc retrieved the 10 most similar images from this subset, so that both pipelines operated on the same retrieved set. HLoc derives the 2D--3D correspondences from the precomputed 3D points, whereas our pipeline triangulates the local model on the fly from the 10 retrieved images.

\begin{table}[ht]
\centering
\caption{Local SfM (ours) versus global SfM (HLoc; \citealp{sarlin_coarse_2019}): percentage of poses localized against the 2025 reference campaign within seven error thresholds per camera, aggregated over the five scenes. Both methods use MegaLoc ($k = 10$), ALIKED at 1024 pixels and a RANSAC threshold of 12 pixels (HLoc default).} \label{tbl_abl_hloc_vs_ours}
\setlength{\tabcolsep}{2pt}
\begin{tabular}{l *{5}{S[table-format=2.1]}*{2}{S[table-format=3.1]}}
\toprule
 \multirow{2}{*}{Camera}& {0.01m}& {0.025m} &{0.05m} &{0.1m}&{0.2m}&{0.5m} &{1m } \\
& {0.1$^\circ$} & {0.25$^\circ$} & {0.5$^\circ$} & {1$^\circ$} & {2$^\circ$} & {5$^\circ$} & {10$^\circ$}\\ \midrule
\multicolumn{8}{c}{\textit{HLoc \citep{sarlin_coarse_2019}}} \\
Canon 24&  6.8&43.5&78.5&96.7&99.9&100.0&100.0 \\
Canon 35& 14.1&52.6&78.5&94.6&98.7&99.5&99.6 \\
GoPro& 9.6&47.6&81.6&95.9&99.5&100.0&100.0 \\
iPhone&  6.7&32.4&61.5&87.2&95.1&97.0&97.4 \\
\midrule
\multicolumn{8}{c}{\textit{Ours}} \\
Canon 24& 4.7&32.8&62.7&91.1&97.6&99.5&99.5 \\
Canon 35& 6.6&28.7&62.7&91.4&97.8&99.1&99.2 \\
GoPro& 7.6&44.2&80.5&96.8&99.8&99.9&99.9 \\
iPhone& 6.1&33.5&60.3&84.9&94.2&96.2&96.6 \\
\bottomrule
\end{tabular}
\end{table}

Consistent with \citet{panek_guide_2026}, HLoc outperforms our pipeline in the high-accuracy regime, but the margin depends strongly on the query camera (Table~\ref{tbl_abl_hloc_vs_ours}). At 0.05\,m / 0.5$^\circ$ it leads by 15.8\,pp for both Canon lenses and by only 1.1\,pp for the GoPro and 1.2\,pp for the iPhone. At 0.025\,m / 0.25$^\circ$ the spread is wider, from 23.9\,pp for the Canon 35\,mm to 3.4\,pp for the GoPro, whereas for the iPhone our pipeline is 1.1\,pp ahead. From 0.1\,m / 1$^\circ$ onward the two pipelines are largely on par, with differences of at most 5.6\,pp and a small advantage for our pipeline with the GoPro images. Per scene and camera, HLoc leads in the accurate category in 15 of the 20 combinations, in all ten for the two Canon lenses but in only five of ten for the GoPro and the iPhone. A precomputed, globally consistent model thus yields more accurate 3D points. This improves the estimated pose only where the 3D point coordinates are the dominant source of pose error. It has little effect where the geometry of the retrieved images or the quality of the query imagery dominates instead.

\section{Discussion}\label{discussion}
\subsection{Accuracy potential of sparse feature-based visual localization}\label{dis_accuracy_potential}

For daytime imagery and pre-calibrated cameras, the best-performing local features achieve median pose errors of 1--5\,cm in translation and 0.03--0.1$^\circ$ in rotation (Tables~\ref{tbl_exp_image_res_feats} and~\ref{tbl_exp_local_scenes_agg_median_pose_error}). Reaching the lower end of that range requires image resolutions of at least 2048 pixels and a suitable local feature representation, while the corresponding absolute accuracy additionally depends on the georeferencing of the scene representation. We first discuss the geometric factors influencing pose accuracy, then the local feature representations and what their differences imply for the evaluation of visual localization, and finally the scene representation and the relation between relative and absolute accuracy.

In PnP-based pose estimation, the accuracy of the resulting pose depends mainly on the angular accuracy with which the 3D points are observed, their distribution in image and object space, and the accuracy of their coordinates. Increasing the angular resolution of the image observations by using high-resolution images substantially improved the accuracy of the final poses (Table~\ref{tbl_exp_image_res_feats} and Figure~\ref{fig_exp_img_res_feat}). Doubling the image resolution from 1024 to 2048 pixels reduces the median pose error by roughly 25 to 45\,\% for most tested local feature representations, and further increases can yield additional, though smaller, improvements depending on the representation. The localization accuracy of the keypoints also matters, as sub-pixel-accurate keypoints yield more accurate poses \citep{lindenberger_pixel-perfect_2021, kim_learning_2025, vultaggio_investigating_2024}, especially at low image resolutions. Among the learned representations, ALIKED and RDD provide sub-pixel keypoint locations through a differentiable keypoint detection module \citep{alike_zhao_2022}. Both lead in the high-accuracy regime. At the strictest threshold, RDD is ahead of all other representations at every tested resolution, while ALIKED matches or exceeds SuperPoint up to 4096 pixels (Table~\ref{tbl_exp_image_res_feats}). SIFT also provides sub-pixel keypoint locations, but its performance is limited by the sensitivity of its handcrafted descriptor to the appearance changes between the campaigns rather than by keypoint precision.

Nevertheless, keypoint precision alone does not explain the performance differences between local feature representations. Among the representations with pixel-level keypoints, DISK matches or exceeds ALIKED at the strictest threshold at 1024 and 2048 pixels, and SuperPoint does so at full resolution. Local features with sub-pixel keypoint localization therefore tend to lead at the strictest thresholds, although other factors also influence the resulting pose accuracy. Following \citet{zhang_reference_2021}, we hypothesize that the quantity and distribution of keypoints across the image plane are equally important, so that many well-distributed but less accurately localized correspondences can outperform a few highly accurate ones, particularly in geometrically unfavorable configurations, although our experiments do not isolate this effect. Wide field-of-view cameras can provide a favorable distribution of image observations, consistent with the lowest median translation errors of the GoPro images (Section~\ref{exp_consumer_query}), although isolating the effect of the field of view would require a controlled comparison with identical trajectories.

Learned feature representations typically degrade on out-of-domain data \citep{bonilla_mismatched_2025}. The resolution-dependent optimum observed for most representations (Table~\ref{tbl_exp_image_res_feats}) may therefore be attributed to the lack of high-resolution training data. SuperPoint \citep{detone_superpoint_2018}, which benefited most consistently from higher resolutions, is trained with synthetic homographic warps rather than SfM-derived correspondences. This may explain why it generalizes to high-resolution images and extracts repeatable, well-localized keypoints even on out-of-domain data. Recent methods such as RDD additionally leverage the global image context to produce more robust descriptors. Tiled feature extraction limits this context to the tile, which may explain why RDD does not improve beyond 2048 pixels at the two strictest thresholds, although it continues to benefit from higher resolution at the coarser ones.

With the consumer-grade query images, RDD is the strongest representation, but no representation is best across all cameras, metrics and scenes (Section~\ref{exp_consumer_query}). DISK and XFeat, which perform comparably to the other representations on standard benchmarks such as Aachen \citep{sattler_benchmarking_2018}, underperform with both reference epochs. This is consistent with \citet{bonilla_mismatched_2025}, who argue that the image matching method should be selected carefully for the specific scenario. A similar accuracy potential was demonstrated by \citet{lindenberger_pixel-perfect_2021}, \citet{zhai_splatloc_2025}, and \citet{ress_3d_2025}, but on datasets with few challenging conditions (ETH3D \citep{schops_multi-view_2017}, 7-Scenes \citep{shotton_scene_2013}, and a proprietary indoor dataset), which emphasizes the importance of robust image matching under challenging conditions.

The performance differences between the local feature representations also show that the datasets currently used to evaluate visual localization give only a limited indication of its accuracy potential. Their ground-truth poses are typically accurate to the decimeter level (Table~\ref{table_dataset_comparison}), so methods cannot be evaluated reliably at fine error thresholds. On the day subset of Aachen Day--Night \citep{sattler_benchmarking_2018}, \citet{chen_rdd_2025} report a spread of only 2.7\,pp between SuperPoint, DISK, ALIKED, XFeat and RDD at the strictest threshold of 0.25\,m and 2$^\circ$. In the accurate category of our experiments, by contrast, XFeat lags behind the best representation by 36\,pp at 1024 pixels (Table~\ref{tbl_exp_image_res_feats}) and, for the consumer cameras, by 32 to 38\,pp and 25 to 33\,pp against the 2023 and 2025 reference data (Tables~\ref{tbl_exp_local_scenes_agg_cat_2023} and~\ref{tbl_exp_local_scenes_agg_cat_2025}). Such differences only become visible with ground-truth poses whose uncertainty is well below the evaluated error thresholds (Section~\ref{m_dataset_query_sensitivity}).

Precomputed, globally consistent SfM models typically provide the most accurate 3D points but require extensive preprocessing, whereas meshes and city models abstract geometric and visual information. Our pipeline avoids both by triangulating the 3D points directly from the georeferenced high-resolution imagery, at a cost of up to 15.8\,pp fewer poses within 0.05\,m and 0.5$^\circ$ compared with a globally consistent SfM model (Table~\ref{tbl_abl_hloc_vs_ours}). We attribute this to the weaker triangulation geometry of the local model, and the camera dependence reported in Section~\ref{exp_local_vs_global} supports this interpretation. The penalty is largest for the two Canon lenses and absent for the GoPro, whose wide field of view already provides a favorable distribution of image observations among the few retrieved images, whereas for the iPhone the pose error is dominated by the query imagery rather than by the accuracy of the 3D points.

The errors reported above exclude the georeferencing error of the reference imagery. Combined with the georeferencing RMSE of 0.025\,m (Section~\ref{m_dataset_query_sequences}), the median relative errors of 1--5\,cm indicate absolute errors of approximately 3--6\,cm, dominated by the georeferencing for the most accurate poses. As this combines medians with an RMSE, the comparison with survey-grade GNSS positioning is indicative rather than strict. Approaching survey-grade accuracy requires both accurate relative localization and an accurately georeferenced scene representation, and is ultimately bounded by the latter.

\subsection{Reference epoch and map maintenance}\label{dis_reference_epoch}
For all cameras and feature representations, localization against the 2025 reference imagery is more accurate than against the 2023 reference imagery, and combining both epochs yields median errors in between (Tables~\ref{tbl_exp_local_scenes_agg_cat_2023}--\ref{tbl_exp_local_scenes_agg_median_pose_error}). This ordering cannot be attributed to a single factor, as the two campaigns differ in several respects at once. The 2025 campaign was acquired approximately one month after the query images, whereas the 2023 campaign preceded them by about 16 months, during which ongoing construction work changed parts of the street environment. The vegetation (trees free of foliage) of the 2025 campaign matches that of the query images, and its illumination (overcast, diffuse light) that of the two query scenes captured under cloudy conditions, whereas the 2023 campaign was acquired under predominantly sunny conditions with shadow patterns and foliage. In addition, the sensor setup changed between the campaigns, including a panorama camera with a substantially higher resolution in 2025 (Table~\ref{mms_sensor_setup}). Our experiments therefore show the combined effect of these factors but cannot separate them.
 
Adding reference data does not necessarily improve localization, as combining both epochs performs worse than using the 2025 imagery alone. We see two plausible explanations, which our experiments do not separate. Both lie in the reference candidate selection, which retrieves a fixed number of the most similar reference images ($k = 10$). First, with both epochs in the database, some of the k retrieved images are from the 2023 campaign, which differ more in appearance from the query images and are likely to yield fewer correct correspondences, also among the reference images. Second, images of both epochs acquired at nearly the same positions may be retrieved together, so that the retrieved set covers fewer distinct viewpoints and the triangulation geometry of the local model becomes weaker. Geometric inconsistencies between the epochs, in contrast, are unlikely to be the cause, as both campaigns are co-registered with a 3D RMSE of 0.005\,m at the independent tie points (Table~\ref{corregistration_results_mm_campaigns_rel}). If confirmed, this would suggest selecting reference candidates not by visual similarity alone, but also by their geometric configuration, e.g., by requiring a minimum pose difference between the selected images.

For the maintenance of image-based scene representations, these results suggest that recent reference imagery should be preferred over accumulated imagery, e.g., by replacing outdated images, by restricting the retrieval to the most recent epoch available in a region, or by map summarization and change-aware selection. Without such measures, the scene representation also grows without bounds. Updating and bounding long-term localization maps has been studied in robotics \citep{burki_map_2018}, and adapting such map management techniques is a necessary step toward consistently high-accuracy visual localization. Due to the lack of accurately co-registered multi-epoch reference data, this has not yet been investigated extensively for high-accuracy visual localization. The two co-registered reference epochs of our dataset provide a basis for such investigations.

\subsection{Applications of accurate visual localization}\label{dis_applications}
In the following, we distinguish between applications that are directly supported by our experiments and prospective applications that our results motivate but do not test. Regarding applications that require accurate pose information with respect to a reference frame, the georeferencing of the scene representation is critical, as it determines the achievable absolute pose accuracy (Section~\ref{dis_accuracy_potential}). In addition, an up-to-date scene representation is needed to maintain consistently high pose accuracy (Section~\ref{dis_reference_epoch}). Reference mapping therefore has to be repeated at regular intervals, and each new epoch requires accurate georeferencing and precise co-registration with the existing epochs. Our experiments directly demonstrate the localization of mobile mapping imagery against an earlier campaign. In the best case, about 86\,\% of the 2025 images were localized with errors below 0.025\,m and 0.25$^\circ$ relative to the 2023 campaign (Table~\ref{tbl_exp_image_res_feats}). Introducing such poses as observations in integrated georeferencing \citep{eugster_integrated_2012} could reduce the need for manually measured GCPs in subsequent epochs, limiting the manual georeferencing effort largely to the initial reference mapping. More generally, visual localization can be seen as a method for the automated generation of control information. This applies to both co-registration and georeferencing, provided that a precisely georeferenced scene representation is available.

A prospective application is the acquisition of 3D geospatial data with consumer devices. Smartphones equipped with LiDAR sensors are emerging as a cost-effective and user-friendly alternative to traditional surveying equipment, but accurate georeferencing currently requires additional GNSS hardware \citep{rehak_accurate_2025} or GCPs determined with traditional surveying methods. Visual localization could provide this georeferencing directly from the captured imagery. Our experiments support this only in part. The iPhone images were localized with median errors of 0.023\,m and 0.053$^\circ$ (Table~\ref{tbl_exp_local_scenes_agg_median_pose_error}), but they were pre-calibrated (Section~\ref{dis_limitations}).

Neither use case necessarily requires real-time processing, as a few localized frames may be sufficient for georeferencing. The runtime of approximately 3.3\,s per image is therefore not critical (Appendix~\ref{appendix_runtime}). The memory needed for the preprocessed global image descriptors and local features is more relevant, particularly when scaling to entire cities or states with millions of images (Appendix~\ref{appendix_memory}).

\subsection{Limitations and future work}\label{dis_limitations}
While our experiments demonstrate the accuracy potential of visual localization, this potential is established under specific conditions that bound the scope of our findings. First, the query images were acquired in five scenes of a single suburban municipality, during daytime in late winter under cloudy and sunny conditions. Our dataset includes only mild weather-induced appearance changes and no imagery acquired under low-light conditions, such as at dusk or during nighttime (Table~\ref{table_dataset_comparison}). The transfer of the accuracy potential to other environments, such as dense urban areas, and to a wider range of seasonal and illumination conditions therefore remains to be shown. Localization against the 2023 reference campaign nevertheless provides a first indication, as it combines an interval of about 16 months with a change of season and illumination and with intervening construction work. Under these conditions the best median translation error per camera rises from between 2.0 and 3.1\,cm to between 3.1 and 5.0\,cm (Table~\ref{tbl_exp_local_scenes_agg_median_pose_error}), a degradation by factors of 1.2 to 2.2. The reference database likewise covers a single 10\,km street network, so the scalability of the pipeline to city- or region-wide databases was not evaluated beyond the memory and runtime figures reported in Appendix~\ref{appendix}. Second, all query images were pre-calibrated and undistorted using the camera calibration estimated during ground-truth generation, and the intrinsics were kept fixed during pose estimation. This is typically given for imagery captured by professional vision-based MMS, but not for consumer-grade devices, whose calibration is usually unknown or less accurate. The effect of inaccurate or unknown intrinsics was not evaluated. Third, the scene representation consists of high-resolution imagery acquired with professional mobile mapping systems and georeferenced with an accuracy of a few centimeters. Consequently, the reported results represent the accuracy potential under favorable conditions and may degrade with less accurate scene representations, strong lens distortions or more challenging acquisition conditions. The contributions of query and reference resolution were not separated, as varying them independently would introduce a scale difference between the two and confound resolution with scale.

Fourth, ground truth and localization pipeline follow the same photogrammetric principle. \citet{brachmann_limits_2021} and \citet{zhang_reference_2021} found that localization methods resembling the ground-truth procedure tend to be favored. In our case, the two differ in software, features, matching and 3D points. No automatic correspondences between query and reference images enter the ground truth, which is connected to the reference imagery exclusively through manually measured points (Section~\ref{m_dataset_query_sequences}). The same applies to the mobile mapping queries in Sections~\ref{exp_reference_candidates} and~\ref{exp_high_res_imagery}, as the two campaigns are connected exclusively via manually measured GCPs and TPs (Section~\ref{m_dataset_reference_imgs}). The two share only the reference image poses and the query camera calibration. The reference poses define the reference frame and are therefore inherent to the relative accuracy reported here. The calibration is estimated in the ground-truth adjustment and used to undistort the query images, so a biased calibration would shift ground-truth and estimated poses in a similar way. The sensitivity analysis in Section~\ref{m_dataset_query_sensitivity}, however, shows no compensation between calibration and image poses and limits such correlated errors to the millimeter level.

Fifth, the evaluation is limited to sparse local features. Dense matchers \citep{edstedt_roma_2024}, approaches built on dense correspondences such as ImLoc \citep{jiang_imloc_2026}, feature refinement \citep{lindenberger_pixel-perfect_2021} and feed-forward reconstruction models \citep{leroy_mast3r_2025} were not evaluated. Most of these methods operate at substantially reduced image resolutions, and their computational and storage costs increase considerably at the resolutions considered here. Whether they can exploit high-resolution imagery to reach comparable or higher accuracy remains to be investigated.

In future research, we intend to address several of these limitations. First, we plan to extend the dataset with additional areas along the road network and more challenging conditions, such as diverse weather, seasonal changes and, in particular, imagery acquired during dusk and nighttime. Second, we aim to investigate the use of uncalibrated imagery from consumer-grade devices for georeferencing, with a focus on robust localization under multiple, simultaneously occurring adverse conditions.

\section{Conclusion}\label{conclusion}
In this paper, we investigated the accuracy potential of sparse feature-based visual localization using high-end street-level imagery. Due to the lack of a suitable dataset for this purpose, we introduced the FHNW Muttenz dataset, which is publicly available to foster future research. It comprises high-resolution imagery from two mobile mapping campaigns of a 10\,km street network and query sequences acquired with four different cameras in five scenes. All images are co-registered to the reference imagery, and the ground-truth poses of the query images have an uncertainty of 6--9\,mm in position and well below 0.1$^\circ$ in orientation. To evaluate the accuracy potential on this dataset, we used a modular visual localization pipeline built from established components. It relies on the simple scene representation of georeferenced imagery and combines a prior-guided reference candidate selection strategy with SfM for building a local model on the fly and subsequently estimating the query image's pose. For daytime imagery and pre-calibrated cameras, our experiments yield median pose errors of 1--5\,cm in translation and 0.03--0.1$^\circ$ in rotation relative to the reference imagery. Furthermore, the experiments revealed performance differences between local feature representations that are not resolved at the coarse error thresholds of existing benchmarks and only become visible with highly accurate ground-truth poses. The experiments further show that localization accuracy depends on the epoch of the reference imagery, and that accumulating imagery from several epochs does not improve it. Maintaining an image-based scene representation therefore requires replacing or selecting reference imagery rather than merely adding to it. The absolute accuracy is additionally bounded by the georeferencing of the scene representation, which is about 2.5\,cm in our case. Given an accurately georeferenced scene representation, visual localization can therefore approach the accuracy of survey-grade GNSS positioning, with the advantage of yielding 6-DoF poses in any environment mapped in advance. This motivates its use as a complementary positioning modality to GNSS and for the automated georeferencing of imagery, which traditionally involves a substantial amount of manual effort. Extending these findings to uncalibrated consumer devices, low-light conditions and other environments remains future work.

\section*{Declarations}


\paragraph{Acknowledgements}
Not applicable

\paragraph{Funding}
This research was supported by the Forschungsfonds Kanton Aargau and our industrial partner iNovitas AG through Grants 20200331\_04\_PReCIS and 20240930\_16\_GRAVIS3D, as well as by Innosuisse via the Innovation Booster Artificial Intelligence program under Grant 03.10.2024-07.

\paragraph{Conflicts of interest/Competing interests}
The authors declare that they have no known competing financial interests or personal relationships that could have appeared to influence the work reported in this paper.

\paragraph{Availability of data and material}
The data used in this work is publicly available at:\newline \url{https://fhnw-muttenz-vl-dataset.github.io/}.

\paragraph{Code availability} The visual localization pipeline is composed of publicly available components. The methods used and all settings deviating from their defaults are given in Section~\ref{exp_implementation_details}, so that the experiments can be reproduced from the published implementations and the released dataset.

\paragraph{Authors' contributions}
\textbf{J.M.:} Conceptualization, Methodology, Validation, Investigation, Data curation, Writing -- original draft, Visualization. \textbf{S.N.:} Conceptualization, Methodology, Writing -- review \& editing, Supervision, Project administration, Funding acquisition. \textbf{P.T.:} Writing -- review \& editing, Resources, Data curation. \textbf{N.H.:} Methodology, Writing -- review \& editing, Supervision.

\appendix

\section{Memory requirements and runtime}\label{appendix}
\subsection{Memory consumption}\label{appendix_memory}

Our visual localization pipeline requires precomputed global descriptors stored in a vector database. For runtime and efficiency reasons, we also use precomputed local features. The memory requirements of both the global image descriptors and the local features may therefore be of interest when selecting suitable methods. This is particularly true when the environment spans entire cities or states. Table~\ref{tbl_abl_memory_ir} reports the memory footprint of a single global image descriptor for each IR method. Analogously, Table~\ref{tbl_abl_memory_lf} presents the memory needed to store 1024 sparse local features per representation. All values reflect uncompressed memory usage to ensure fair and consistent comparison.

\begin{table}[ht]
\centering
\caption{Memory requirement of one global descriptor per IR method.}\label{tbl_abl_memory_ir}
\begin{tabular}{l S[table-format=5.0] c S[table-format=3.0]}
\toprule
 Method & {Desc dim}& {dtype} & {Data [KB]}\\ \midrule
 NetVLAD  & 4096 & float32 & 16 \\ 
 CosPlace & \centering{512 {\textbar} 2048} & float32 & \centering{2 {\textbar} 8} \\
 EigenPlaces & \centering{512 {\textbar} 2048} & float32 & \centering{2 {\textbar} 8} \\
 AnyLoc & 49152 & float32 & 192\\
 DINOv2 SALAD & 8448& float32 & 33 \\
 MegaLoc & 8448 & float32 & 33 \\
 \bottomrule
\end{tabular}
\end{table}

\begin{table}[ht]
\centering
\caption{Memory requirement to store 1024 sparse local features per representation.}\label{tbl_abl_memory_lf}
\setlength{\tabcolsep}{3.5pt}
\begin{tabular}{l S[table-format=3.0] cc S[table-format=4.0]}
\toprule
 \multirow{2}{*}{Method} & \multicolumn{2}{c}{Descriptor} & {Keypoint}& {Data per 1024} \\
 & {length} & dtype & dtype & {features [KB]} \\ \midrule
 SuperPoint& 256 & float32 & int16 & 1028 \\ 
 DISK& 128 & float32 & int16 & 516\\ 
 ALIKED & 128 & float32 &float32 & 520  \\ 
 XFeat& 64 & float32 & int16 & 260\\ 
 RDD & 256 & float32 & float32 & 1032\\ 
 \bottomrule
\end{tabular}
\end{table}

\subsection{Runtime}\label{appendix_runtime}
The runtime evaluation was conducted on a workstation equipped with an AMD Ryzen Threadripper 7970X 32-Core @ 4.0 GHz CPU and an NVIDIA RTX 4090 GPU. The runtime of our visual localization pipeline strongly depends on the chosen settings, including the local feature representation, image resolution, and the number of features extracted. We provide the runtime of each major pipeline step and its outputs with respect to time-relevant settings.

We randomly selected 1000 images from the front-facing monocular camera of the 2025 campaign as a test set. These images remained unchanged across all runtime evaluations. We applied our pipeline with settings identical to those used in the experiment investigating the role of high-resolution imagery (Section~\ref{exp_high_res_imagery}) and measured the time.

Tables~\ref{tbl_abl_runtime_descs},~\ref{tbl_abl_runtime_lf}, and~\ref{tbl_abl_runtime_im} report the average runtime for extracting global image descriptors, sparse local features, and image matching, respectively. Table~\ref{tbl_abl_runtime_ir} reports the average runtime for our local SfM-based pose estimation procedure. In addition to the total runtime (Total), we also report sub-part runtimes, such as geometric verification (G), point triangulation from known poses and camera intrinsics (T), and pose estimation via PnP+RANSAC followed by non-linear pose refinement (PnP) (Table~\ref{tbl_abl_runtime_ir}).

\begin{table}[htb]
\centering
\caption{Average runtime [ms] for extracting one global descriptor per IR method.}\label{tbl_abl_runtime_descs}
\begin{tabular}{l S[table-format=2.1]}
\toprule
 Method & {Time [ms]} \\ \midrule
 NetVLAD  & 51.4  \\ 
 CosPlace & 50.6 \\ 
 EigenPlaces & 50.3 \\
 AnyLoc  & 101.6 \\
 DINOv2 SALAD & 56.7  \\
 MegaLoc &  56.2 \\
  \bottomrule
\end{tabular}
\end{table}

\begin{table}[htb]
\centering
\caption{Average runtime [ms] of the tiled feature extraction for one image per resolution and local feature (S: SuperPoint, D: DISK, A: ALIKED, X: XFeat, R: RDD).}\label{tbl_abl_runtime_lf}
\begin{tabular}{c S[table-format=3.1]S[table-format=3.1]S[table-format=3.1]S[table-format=3.1]S[table-format=4.1]}
\toprule
  Image res & {S} & {D} & {A} & {X} & {R}  \\ \midrule
  1024& 75.6 & 97.9&   85.4 & 80.5  & 196.5  \\ 
  2048& 122.4& 181.8&  137.8& 117.8 & 644.8  \\ 
  4096& 243.3& 451.0&  269.0& 202.1 & 2347.0 \\ 
  full& 318.7& 698.6&  373.6& 249.6 & 4114.5 \\ 
  \bottomrule
\end{tabular}
\end{table}

\begin{table}[htb]
\centering
\caption{Average matching runtime [ms] per number of features for one image pair (P) and for exhaustive matching of 11 images, i.e., 55 pairs (EM). Matchers: LightGlue \citep{lindenberger_lightglue_2023}; LighterGlue \citep{potje_xfeat_2024} for XFeat.}\label{tbl_abl_runtime_im}
\begin{tabular}{r S[table-format=3.1] S[table-format=4.1] S[table-format=2.1] S[table-format=4.1]}
\toprule
\multirow{2}{*}{\# Features} & \multicolumn{2}{c}{LightGlue [ms]} & \multicolumn{2}{c}{LighterGlue [ms]} \\ 
& {P} & {EM} & {P} & {EM} \\ \midrule
1024 & 8.7&    476.3   &5.7  &  315.7\\
2048 & 10.7&   586.9   &6.9  & 379.5\\
4096 & 19.5&   1071.4  &7.8  & 429.6\\
6144 & 34.5&   1899.7  &11.1 & 612.2 \\
8192 & 45.8&   2520.1  &18.1 & 996.6 \\
10240 & 70.6&  3885.2  &23.9 & 1314.5 \\
12288 & 89.0&  4893.4  &29.1 & 1603.3 \\
14336 & 120.2&  6613.8 &36.3 & 1997.6 \\
  \bottomrule
\end{tabular}
\end{table}

\begin{table}[htb]
\centering
\caption{Average runtime [ms] of the local SfM-based pose estimation with 10 reference images per number of features. Geometric verification (G), triangulation (T), PnP-based pose estimation (PnP) and total. Precomputed SuperPoint features, matched with LightGlue.}\label{tbl_abl_runtime_ir}
\begin{tabular}{S[table-format=5.0] S[table-format=3.1]  S[table-format=4.1] S[table-format=4.1] S[table-format=4.1]}
\toprule
 {\# Features}& {G} & {T} & {PnP} & {Total} \\ \midrule
1024&143.4&140.8&55.3&388.3    \\
2048&170.4&269.6&82.6&576.5   \\
4096&261.2&517.3&130.0&974.0   \\
6144&315.0&639.1&149.8&1178.3   \\
8192&362.6&756.1&164.5&1365.3   \\
10240&412.5&942.9&188.9&1634.7   \\
12288&455.2&1111.6&207.9&1872.6   \\
14336&486.3&1242.9&220.7&2056.7   \\
  \bottomrule
\end{tabular}
\end{table}

To obtain the approximate runtime of the entire pipeline, the runtimes in Tables~\ref{tbl_abl_runtime_descs},~\ref{tbl_abl_runtime_lf},~\ref{tbl_abl_runtime_im}, and~\ref{tbl_abl_runtime_ir} for the desired methods must be summed. For example, using MegaLoc as the IR method and ALIKED as the local feature representation with 6144 features at an image resolution of 2048, our pipeline takes approximately 3.3 seconds to localize a query image. Note that loading the query image and the preprocessed local features as well as querying the vector database are neglected in the runtime ablation study. Querying the vector database typically takes 10--20 ms and is therefore negligible. Loading times are also excluded, as the preprocessed features were stored on external drives in our setup, and the resulting I/O cost is not representative of an optimized deployment.

\bibliographystyle{spbasic}      
\bibliography{bibliography}   

\end{document}